%% file: main.tex
\documentclass{article}

\usepackage{hyperref}

\usepackage[utf8]{inputenc}
\usepackage[english]{babel}
\usepackage{amsmath,amsthm,amsfonts,amssymb,mathrsfs}
\usepackage{comment,color,enumerate,graphicx}
\usepackage{hyperref,pdfpages,proof,url}
\usepackage{relsize,stmaryrd,tikz}
\usepackage{multirow,dsfont}
\usepackage{comment} 

\input{commandes}

\begin{document}

\title{A Computational Model for Logical Analysis of Data}
\author{
Danièle Gardy\footnote{David Laboratory, Université de Versailles Saint-Quentin-en-Yvelines, France}
\and
Frédéric Lardeux, Frédéric Saubion\footnote{LERIA, Université d'Angers, France}}

\maketitle

\begin{abstract}

Initially introduced by Peter Hammer, Logical Analysis of Data (LAD) is a methodology that aims at computing a logical justification for dividing a group of data in two groups of observations, usually called the positive and negative groups. 
Let us consider this partition into positive and negative groups as the description of a partially defined Boolean function; the data is then processed to identify a subset of attributes, whose values may be used to characterize the observations of the positive groups against those of the negative group. 

LAD constitutes an interesting rule-based learning alternative to classic statistical learning techniques and has many practical applications. 
Nevertheless, the computation of group characterization may be costly, depending on the properties of the data instances. 
A major 	aim of our work is to provide effective tools for speeding up the computations, by computing some \emph{a priori} probability that a given set of attributes does characterize the positive and negative groups.
To this effect, we propose several models for representing the data set of observations, according to the information we have on it.
These models, and the probabilities they allow us to compute, are also helpful for quickly assessing some properties of the real data at hand; furthermore they may help us to better analyze and understand the computational difficulties encountered by solving methods.  
 
Once our models have been established, the mathematical tools for computing probabilities come from Analytic Combinatorics.
They allow us to express the desired probabilities as ratios of generating functions coefficients, which then provide a quick computation of their numerical values.
A further, long-range goal of this paper is to show that the methods of Analytic Combinatorics can help in analyzing the performance of various algorithms in LAD and related fields.

\end{abstract}



\newpage

\section{Introduction}
\label{sec:intro}

Understanding data and their relationships is an important challenge in artificial intelligence that can be solved by using various machine learning and data mining techniques. 
Logical analysis of data (LAD)~\cite{hammer1986,Crama1988,Hammer2006,Chikalov2013} combines concepts from partially defined Boolean functions and optimization in order to characterize sets of data. 
LAD is an interesting rule-based learning alternative to classic statistical machine learning techniques (e.g., clustering algorithms). 
Searching for some kind of patterns, LAD shares some similarities with the frequent itemset mining task in association rules extraction \cite{agrawal1993mining}. 
Nevertheless, the purpose and hypotheses are rather different in both approaches, and LAD targets other purposes. 
Hence, LAD is now recognized as an important rule-based learning technique with many potential applications (see \cite{LejeuneLLRY19} for a recent survey).

\paragraph{Basic Concepts and Motivations}

In order to illustrate the main motivations of our work, we begin with a simple case that clearly states the main concepts and purposes of LAD methodology.
Let $P$ and $N$ be two sets of observations (respectively positive and negative observations, we call them groups) defined over a set of Boolean attributes. 
LAD aims at finding a logical explanation justifying that some observations belong to $P$ and the other ones to $N$.
Consider the initial data as a partially defined Boolean function \cite{ibaraki_crama_hammer_2011} with regards to $P$ and $N$; it is then possible to compute a Boolean formula that evaluates to \True\ for observations in $P$ and to \False\ for observations in $N$. 
The underlying optimization problem consists in finding the most convenient representation of this Boolean function, and the problem may be reformulated as {\em the selection of the most suitable subset of Boolean attributes as regards the properties of the expected representation}.
Note that LAD differs from statistical learning techniques that aim at selecting attributes whose values are somehow correlated to the positive and negative groups of observations.

In its original formulation, LAD methodology focuses on the computation of \emph{patterns} that cover the positive examples. 
These patterns are intended to be clearly understood by human experts, as explanations learned from the partial knowledge available. 
But sometimes the human experts need short explanations (e.g., to limit the number of tests involved in a diagnosis); in such a case the logical analysis is related to the \emph{minimization of Boolean functions}~\cite{McCluskey1956}.  

A typical application of LAD is the diagnosis of biological/medical data to help a physician identify common characteristics that are shared by a group of sick patients, as opposed to a group of healthy patients. 
Apart from biology and medicine (\cite{Kholodovych2004,Alexe2005,Alexe2006a}), LAD has also been applied to many practical application domains such as engineering \cite{Bennane2012,MortadaCYL12,MortadaYL14}, transportation \cite{Dupuis2012}, finance \cite{HammerKL12,KoganL14}, or healthcare \cite{JocelynCOY17}.

\begin{example}
\label{example:intro}
Let us consider the simple illustrative example given in Table~\ref{table-exemple-intro}.
The data consists of $7$ observations on $8$ Boolean attributes (labeled from a to h; as usual we identify 0 with \False\ and 1 with \True) and the observations are partitioned in two groups~$P$ and~$N$.

\medskip
\begin{table}[h]
\begin{center}
\begin{tabular}{|c|c|c|c|c|c|c|c|c|c|}
\hline
\multirow{2}{*}{Observations}&\multirow{2}{*}{Groups} & \multicolumn{8}{c|}{Attributes}\\
\cline{3-10}
&&a & b & c & d & e & f & g & h\\
\hline
1&\multirow{3}{*}{P} & 0&1&0&1&0&1&1&0 \\
2& & 1&1&0&1&1&0&0&1\\
3& & 0&1&1&0&1&0&0&1 \\
\hline
4& \multirow{4}{*}{N} & 1&0&1&0&1&0&1&1 \\
5& & 0&0&0&1&1&1&0&0\\
 6& & 1&1&0&1&0&1&0&1 \\
 7& & 0&0&1&0&1&0&1&0\\
\hline
\end{tabular}
\caption{\label{table-exemple-intro} Running example with 7 observations and 8 attributes.}
\end{center}
\end{table}

\medskip
\noindent
We wish to characterize the groups $P$ and $N$ by finding a subset of attributes that allows us to differentiate these two groups.
Of course, there may be several such subsets; we then choose one solution according to criteria such as the subset size or the ability to obtain a formula in a specific form, e.g., a normal conjunctive or disjunctive form.
\begin{itemize}
\item
The only attribute that takes the same value, namely~1, on all the observations of~$P$ is~$b$, but this attribute alone cannot be used to explain $P$: the observation $6$ in $N$ has also the value~$1$ for~$b$. 

\item
Consider now combinations of attributes.
The two attributes $f$ and $g$ together are sufficient to characterize $P$ and $N$: they take the same value on observations in~$P$ and opposite values on observations in~$N$.
Let us consider the groups $P$ and $N$ as the partial truth table of a (possibly partial) Boolean function. 
The Boolean formula $(f \wedge g) \vee (\neg f \wedge \neg g)$ is true for the Boolean assignments of $P$ and false for those of $N$; this formula is in disjunctive normal form.
Equivalent formulae are $ (f \vee \neg g) \wedge (\neg f \vee g)$ (in conjunctive normal form) or $f \leftrightarrow g$.
Note that $f \leftrightarrow g$ does not mean that the attributes $f$ and $g$ are equivalent, but simply that the observations for which they take the same value are exactly those of group~$P$. 
Of course, this last formulation may be more understandable for practitioners who want to perform diagnosis by observing whether attributes $f$ and $g$ have the same value or not.
 
\end{itemize}

\end{example}

We emphasize that a formula should be the most convenient for the practitioner/expert, either by minimizing the number of attributes (e.g., to simplify their practical implementation in diagnosis routines) or by minimizing the size of the formula (e.g., to improve their readability).

\medskip
Another way to characterize the observations, and indeed a key concept in LAD methodology, is to identify patterns in observations. 
Informally, a pattern is a set of attributes and values for these attributes, or equivalently a Boolean expression on these attributes, that uniquely characterizes all or some of the positive observations (see Definition~\ref{def:pattern} below for a formal definition of pattern).

\vskip .2cm \noindent
{\bf Example~\ref{example:intro} (continued).}
{\em
E.g.,  $\{ a=0, \, b=1 \}$, i.e. $\neg a \vee b $, is a pattern that is shared by observations $1$ and $3$ but such that no observation in $N$ has this pattern. 
This pattern ``covers'' two observations of $P$, and could be interpreted as a partial explanation of the structure of $P$. 
}

\medskip
Among the sets of patterns, a compromise must be achieved between their number of attributes and their \emph{covering} (number of observations in~$P$) that they provide. 
Concerning the number of attributes of the patterns (which may appear as positive or negative literals from Boolean formula's point of view, some properties have been exhibited to focus on the most relevant ones \cite{HammerKSS04}. 
In particular, \emph{prime} patterns are patterns whose number of attributes cannot be reduced; they are an answer to the simplicity requirement (in terms of attributes). 
A prime pattern is thus the smallest in terms of attributes.
\emph{Strong} patterns correspond to the evidential preference where a larger cover (in terms of observations) is preferred. 
A pattern is indeed strong if there do not exist another pattern whose covering is larger. Therefore a strong pattern is the largest in terms of covered observations.

\medskip
The LAD methodology can be extended to multiple groups of data, as multi-class LAD \cite{KimC15a}.
It has also been studied as a multiple characterization problem (MCP) \cite{Chhel2012}: this problem consists in minimizing the number of attributes necessary to discriminate mutually several groups of observations, instead of considering only two groups. 

Of course, some practical questions must be addressed before processing data with a LAD methodology, such as the representation of the observations as binary vectors \cite{boros1997}, or attribute set reduction. 
From the point of view of solution computation, linear programming models have been proposed \cite{Hammer2006} or, e.g.,  dedicated Branch and Bound based algorithms \cite{chambon2019,ChambonBLS18}. 
An implementation of the methodology is also publicly available\footnote{https://github.com/vauxgomes/lad-weka}.

\paragraph{Contributions}
Our own experience in studying real cases in biology  \cite{Chhel2013,ChambonLSB19a} has highlighted that instances may have various structural hidden properties, and that these properties have a direct impact on the performances of  algorithms and on their ability to reduce the number of attributes necessary for characterizing groups of data. 

In the present paper we investigate several mathematical models to compute some properties of instances. 
These models can be used in various contexts of LAD, either to analyze instances before processing them, or for classifying instances according to some expected specificity. 
E.g., we may wish to compute possible expected values for bounding the size of the required set of attributes within the characterization process. 
Assuming a uniform probability distribution of the values, we wish check whether the actual data fits, or not, this model.
Such a check is useful for improving the computational process, as well as for understanding the inner properties of the initial data groups.

\paragraph{Example of result}
We illustrate the kind of information that can be computed through a mathematical modeling, by using it on real data from biology; what follows gives an example of the kind of results that our approach allows us to obtain.

Let us consider the database used in Section~\ref{sec:base-observations}. 
The set $\Y$ is the minimal set of attributes required to characterize the set~$P$ positive observations against the set~$N$ of negative ones. 
Hence the chosen value for $|\Y|$ corresponds to the minimal number of attributes on real cases. 
We are interested in evaluating the probability that the values of the positive observations projected on these attributes lead to a unique pattern. 
This illustrates the two characterization concepts described in Example \ref{example:intro}. 

Under the assumption that the attributes are independent and that the observations are independent and uniformly distributed among the set of all possible values, the probability that a set $\Y$ of attributes defines a pattern is given by Table~\ref{table:intro-resultat}. 

\begin{table}
\begin{center}
\begin{tabular}{|r||r|r|l|l|l|l|}
\hline 
Instance &  $n_1$ & $n_2$ & $|\Y|$ & $ |\Z|$ & Pr($\Y$ pattern)  & Real \# patterns \\
\hline 
\emph{rch8} & 5 & 127 & 3 & 34  &  \textbf{1} - $4.3 \, 10^{-7}$  & 1\\ 
\emph{ra\_rep1} &  38 & 74 & 12 & 143 &  $4.295 \; 10^{-134}$ & 27\\  
\emph{ra\_rep2} &  37 & 75 & 11 & 62 &  $2.316 \; 10^{-119}$  & 25 \\ 
\emph{ralsto} &  27 & 46 & 5 & 18 & $6.765 \; 10^{-28}$  & 7\\ 
\emph{ra100\_phv} &  21 & 80 & 2 & 48 & \textbf{1} - $2.6 \, 10^{-8}$ & 1 \\ 
\emph{ra100\_phy} &  31 & 74 & 3 & 48 & $6.666 \; 10^{-6}$ & 5 \\  
\emph{ra\_phv} & 22 & 86 & 2 & 68 & \textbf{1} - $4.5 \, 10^{-9}$ & 1 \\ 
\emph{ra\_phy} & 31 & 81 & 3 & 70 & $2.881 \; 10^{-5}$ & 5 \\ 
\hline 
\end{tabular} 
\vskip .2cm
\caption{\label{table:intro-resultat} 
Probability that $\Y$ is a pattern, i.e., that $|\pi_\Y (P)| =1$. 
The first columns present the data (see Section~\ref{sec:base-observations}) :
the numbers $n_1$ and $n_2$ are respectively the sizes of the positive ($P$) and negative ($N$) groups; 
$|\Y|$ and $|\Z|$ are the cardinalities of the attribute subsets.
Real \# patterns is the number of patterns necessary to fully cover $P$ using only attributes of $\Y$, and is included for comparison purposes. }
\end{center}
\end{table}

The numerical values show that for our data the probability that $\Y$ defines a pattern is always close to either 0 or 1.
The probabilities close to 1 correspond exactly to the instances for which the number of attributes necessary to get a single prime pattern coincides with the minimal characterization. 
In this case, the group $P$ shows a unique pattern when projected on this minimal set of attributes. 
The resulting Boolean formula has a single assignment to satisfy. 
The result of the probabilistic computation is thus particularly interesting, as it may lead us to estimate a suitable bound for our algorithms concerning the size of these minimal sets of attributes. 
Conversely, a probability close to $0$ means that several patterns are necessary to cover $P$. 
We observe that the values of these probabilities are coherent with the number of patterns that are induced by a choice of $\Y$: the lower the probability, the higher the number of patterns with attributes from in $\Y$.
This gives an indication of how our probabilistic estimations may be used in algorithms to focus on relevant bounds for  interesting pattern sizes.

\paragraph{Mathematical tools}
As our aim is to compute probabilities for various situations relative to an instance, i.e., to a global group of observations, we first have to define a probability distribution on the instance and its attributes; this is done in Section~\ref{sec:hypotheses-probabilistes}.
In the frequent case of uniform distributions, which is the only case we consider in this paper, computing the probability of a subset~$B$ of a larger set~$A$ amounts to an enumeration problem: the probability of~$B$ is obtained as the size of~$B$ divided by the size of~$A$.
Powerful tools for enumerating sets of various types have been established through generating functions, most notably in Analytic Combinatorics, and we make heavy use of these tools to compute the various probabilities we are interested in, as ratios of well-chosen generating function coefficients.
Should we wish to consider non-uniform distributions, Analytic Combinatorics technics do allow us to write down the relevant generating functions in various cases, for which we can also consider computations of probabilities.
As regards the actual numerical computations, the use of a Computer Algebra System (here Maple) provides us with a quick computation of the desired probabilities.

\paragraph{Overview of the paper}
We begin by recalling in Section \ref{sec:LAD} the useful material for understanding LAD methodology and the problems we consider in this article.
We then present in~Section \ref{sec:probmod} an overview of the probabilistic models that we  propose for LAD:
 Section \ref{sec:models} gives a detailed description of the different models that can be defined according to which data parameters are known, and Section \ref{sec:overview} introduces the main questions that can be addressed through our formal framework. 
In order to illustrate the possibilities offered by our models, we next present in Section \ref{sec:compex} several numerical computations based on real data, then discuss the scope of our approach together with possible extensions in Section~\ref{sec:scope}.
An appendix (see Section~\ref{sec:appendix}) introduces the reader to the basics of Analytic Combinatorics and covers the mathematical computations needed to establish our numerical results.

\section{Logical Analysis of Data: Background} 
\label{sec:LAD}

In this section, we recall existing work on LAD as regards the two complementary points of view that we have already sketched in the Introduction: minimal characterization in terms of attributes and pattern computation.

	\subsection{Patterns}
	\label{sec:patterns}

The seminal work on LAD \cite{Crama1988} focused on the notion of pattern for two groups $P$ and $N$ of positive and negative observations. 
A pattern aims at identifying a set of attributes that have identical values for several observations in~$P$: of course a pattern must not appear in any observation of~$N$. 
In the presence of more than two groups, we consider that $N$ is the union of all groups except the group considered here as positive ($P$).

\begin{definition}
A {\bf Boolean function} $f$ of $n$ variables, $n\in\mathds{N}$,  is a mapping $f: \mathds{B}^n\mapsto\mathds{B}$, where $\mathds{B}$ is the set $\{0,1\}$. 
A vector $x\in\mathds{B}^n$ is a \textit{\True\ vector} (resp. \textit{\False\ vector}) of the Boolean function $f$ if $f(x)=1$ (resp. $f(x)=0$), and $T(f)$ (resp. $F(f)$) is the set of \textit{\True\ vectors} (resp. \textit{\False\ vectors}) of a Boolean function~$f$. 

A {\bf partially defined Boolean function (pdBf)} on $\mathds{B}^n$ is a pair ($P$,$N$) such that $P$,$N\subseteq\mathds{B}^n$ and $P\cap N=\emptyset$. $P$ is thus the group of positive vectors, and $N$ the group of negative vectors of the pdBf $(P,N)$. 
The notion of partially defined Boolean function is generalized by the following notion of term proposed in (\cite{Hammer2006,Boros2011}). 

A {\bf term} is a Boolean function $t_{\sigma^+,\sigma^-}$ for some groups $\sigma^+,\sigma^-\subseteq\{1,2,...,n\}$ such that $\sigma^+\cap \sigma^-=\emptyset$, whose \True\ set $T(t_{\sigma^+,\sigma^-})$ is of the form  
\[
T(t_{\sigma^+,\sigma^-}) = 
\{
x\in\mathds{B}^n |\text{ }\forall i\in \sigma^+,  x_i=1 \wedge \text{ }\forall j\in \sigma^-,  x_j=0 
\}.
\] 
\end{definition}

A term $t_{\sigma^+,\sigma^-}$ can be represented by a Boolean formula of the form: $t_{\sigma^+,\sigma^-}(x)=(\bigwedge_{i\in \sigma^+}x_i)\wedge (\bigwedge_{j\in \sigma^-}\neg{x}_j)$.

\begin{definition}
\label{def:pattern}
A {\bf pattern} of a pdBf $($P$,$N$)$ is a  term $t_{\sigma^+,\sigma^-}$  such that $|P\cap T(t_{\sigma^+,\sigma^-})|>0$ and $|N\cap T(t_{\sigma^+,\sigma^-})|=0$.
\\
Given a term $t$, the set of attributes (also called variables) defining the term~is
	\[ Var(t_{\sigma^+,\sigma^-})=\{x_i|i\in \sigma^+\cup \sigma^-\}, \]
and the set of literals (i.e. a logic variable or its complement) in $t_{\sigma^+,\sigma^-}$ is
	\[ Lit(t_{\sigma^+,\sigma^-})=\{x_i\cup\bar{x}_j|i\in \sigma^+,\text{ }j\in \sigma^-\} .\]
Given a pattern $p$, we define $Cov(p)=P\cap T(p)$ and say that this group is {\bf covered} by the pattern $p$.
\end{definition}

The group $Cov(p)$ is composed of positive observations such that the pattern~$p$ takes the value \True\ on them, but $p$ can also be true on some negative observations.

\vskip .3cm \noindent
{\bf Exemple \ref{example:intro} (cont.)}
{\em
We illustrate the notion of pattern on our running example given in Section~\ref{sec:intro} and for two groups (see Table~\ref{table-exemple-intro}). 

The formulae $p_1=\neg{a}\wedge b$ and $p_2=\neg{f}\wedge\neg{g}$ are two patterns covering respectively the group of  observations $1$ and $3$ ($p_1$), and the group of observations $2$ and $3$ ($p_2$).

Let us now consider $p_3=f\wedge g$. 
The patterns $p_2$ and $p_3$ use identical attributes:
$Var(p_2)=Var(p_3)$, but $Lit(p_2)\neq Lit(p_3)$. 
The pattern $p_2 \vee p_3$ covers the positive group (since $Cov(p_2)\cup Cov(p_3)=P)$ with only two attributes.
}

	\subsection{The Multiple Characterization Problem (MCP)}
\label{sec:MCP}

\begin{definition}
An {\bf instance} of the MCP is a tuple $(\Omega,{\X},D,G)$ defined by a group of observations $\Omega$, whose elements are data expressed over a set of Boolean attributes ${\cal X}$ encoded by a Boolean matrix of data $D_{|{\Omega}|\times |{\cal X}|}$ and a function $G: \Omega \rightarrow \mathds{N}$, such that  $G(o)$ is the group assigned to the observation $o \in \Omega$.
\end{definition}

The data matrix $D$ is defined as follows:
\begin{itemize}
\item the value $D{[o,a]}$ represents the presence/absence of the attribute $a$ in the observation $o$.
\item a line $D[o,.]$ represents thus the Boolean vector of presence/absence of the different attributes in the observation $o$.
\item a column $D{[.,a]}$ represents thus the Boolean vector of presence/absence of the attribute $a$ in all the observations. 
\end{itemize}

\noindent
{\bf Example \ref{example:intro} (cont.)} 
\emph{
Let us illustrate the previous definitions on our running example database (see Table~\ref{table-exemple-intro}).
The group of observations is 
$\Omega=\{ o_i, i = 1 \ldots 7 \} $
and the function $G: \Omega \to \mathds{N}$ is  fully defined by: 
\begin{eqnarray*}
&& G(o_1) = G(o_2) = G(o_3) =  1;
\\
&& G(o_4) =  G(o_5) =  G(o_6) =  G(o_7) = 2.
\end{eqnarray*}
We build the data matrix $D$ from Table~\ref{table-exemple-intro}, with $D[o_1,.] = 01010110$ and so on, to obtain
\begin{center}
\[
D = \left(
\begin{array}{ c c c c c c c c  }
 0&1&0&1&0&1&1&0 \\
1&1&0&1&1&0&0&1\\
0&1&1&0&1&0&0&1 \\
1&0&1&0&1&0&1&1 \\
 0&0&0&1&1&1&0&0\\
 1&1&0&1&0&1&0&1 \\
 0&0&1&0&1&0&1&0\\
\end{array}
\right)
\]
\end{center}
}

\vskip .2cm
We denote by $G_i$ the group $\{o \in \Omega|G(o)=i\}= G^{-1}(i)$ and by $G$ the union of these groups, i.e. $\Omega$ (we identify a group assignment to its elements).  
In our running example, the groups $G_1$ and $G_2$ are respectively~$P$ and~$N$.

In the following we are only interested in satisfiable MCP, i.e. such that $D$ does not contain two identical observations in two different groups (see~\cite{Chhel2013}). Note that identical lines are deleted (even if two identical lines may correspond  indeed to two different observations on real data, we reduce the matrix for computational purposes).

\begin{property}
\label{def:sat}
A MCP instance $(\Omega,{\cal X},D,G)$ is satisfiable iff:\\

$\nexists (o,o') \in \Omega^2$ such that $o \neq o'$, $D[o,.] = D[o',.]$ and $G(o) \neq G(o')$
\end{property}

We use $D^{\Y}$ for the data matrix reduced to the subset of attributes $\Y \subset {\X}$. 
Given a subset of attributes $\Y \subseteq \X$ and an instance with a function~$G$, let $\pi_\Y (G)$ denotes the projection of the observations of the instance on the attributes of $\Y$. 
Solving  an instance $(\Omega,{\cal X},D,G)$ consists in finding a subset of attributes $\Y \subseteq \X$ such that the projections $\pi_\Y (G_i)$ and $\pi_\Y (G\setminus G_i)$ have no common element, i.e., $(\Omega,S,D^\Y,G)$ is satisfiable.

\begin{definition}
Given an instance $(\Omega,\X,D,G)$, a subset of attributes $\Y \subseteq \X$ is a {\bf solution} iff $\forall (o,o')\in \Omega^2, G(o)\neq G(o') \rightarrow D^{\Y}{[o,.]}\neq D^{\Y}{[o',.]}$. In this case, the matrix $D^{Y}$ is called a solution matrix.
\label{defsolmd}
\end{definition}

An instance of the MCP may have several solutions of different sizes. 
It is therefore important to define a (partial) ordering on solutions in order to compare and classify them. 
In particular, adding an attribute to a given solution $\Y$ generates a new solution $\Y'\supset \Y$; in this case we say  that $\Y'$ is dominated by $\Y$. 

\begin{definition}
A solution $\Y$ is {\bf non-dominated} iff $\forall s\in \Y$, $\exists (o,o')\in{\Omega}^2$ such that  $G(o)\neq G(o')$ and $D^{\Y\backslash \{s\}}{[o,.]}=D^{\Y\backslash \{s\}}{[o',.]}$.
\label{defsolndmd}
\end{definition}

Among the solutions, we are interested in computing solutions of minimal size, as regards the number of attributes they involve. 

\begin{definition}
For a set $\Y$ of attributes, define $|\Y|$ as the number of its attributes.
A solution $\Y$ is {\bf minimal} iff $\nexists \, {\Y'}$ with $|\Y'|<|\Y|$ s.t. $\Y'$ is a solution.
\label{def:minsol}
\end{definition}
According to our notion of dominance between solutions, a minimal solution in not dominated by any other solutions. Intuitively, a minimal (non dominated) solution cannot be reduced unless two identical lines appear in two different groups (and consequently the reduced set of attributes is not a solution).  

\medskip
Through the properties on solutions defined just above, several computational problems can be handled: one may compute one solution, or the set of all possible solutions, according to the dominance relation \cite{ChambonBLS18}. 
In this paper concerned with MCP solving, we focus on two groups of data and on the computation of a minimal solution.


\section{A Probabilistic Model for LAD}
\label{sec:probmod}

\subsection{Notations and Probabilistic Assumptions}
\label{sec:hypotheses-probabilistes}

Consider the set $\X$ of all attributes, and define a partition of~$\X$~: $\Y$ is  the set of attributes\footnote{In the rest of this paper we shall sometimes consider $\Y$ and $\Z$ as attributes, although they actually are sets of attributes, i.e., \emph{aggregated} attributes.} that we want to test as a possible pattern or MCP solution, and its complement $\Z = \X \setminus \Y$ is the set of deleted attributes. 

The respective sizes $d_\Y$ and $d_\Z$ of the domains of the two aggregated attributes are the product of the sizes of the domains for the Boolean attributes in each subset~: $d_\Y= 2^{|\Y|}$ and similarly for $d_\Z$; of course $d_\X = d_\Y \, d_\Z$.

We now make explicit our probabilistic assumptions on an instance, i.e., on attributes and observations.
\begin{itemize}

\item On attributes: we assume that they are independent, i.e., for a random observation no value of a subset of attributes depends on its value on another, disjoint subset of attributes.
We also assume uniformity on the possible values for an attribute, i.e., the values \True\ and \False\ are equally likely.
It is important to notice here that, as we are dealing with Boolean attributes, what matters is not the exact attributes that belong to the subset~$\Y$, but their number, i.e., the size~$|\Y|$.

\item On observations: they are independently drawn without replacement, i.e., all the observations are distinct and the probability of drawing an observation does not depend on the other observations in the instance.

\end{itemize}

Table~\ref{tab:probability-assumptions} sums up our probabilistic assumptions.
\begin{table}[h]
    \centering
    \begin{tabular}{|c|}
    \hline
        An instance has two (aggregated) attributes $\Y$ and~$\Z$ \\
        The attributes take their values in finite domains $D_\Y$ and $D_\Z$ \\
        The distributions on $D_\Y$ and $D_\Z$ are uniform and independent \\
        The observations are drawn independently and without repetition \\
    \hline
    \end{tabular}
    \caption{Probabilistic assumptions on the instances and observations.}
    \label{tab:probability-assumptions}
\end{table}

\subsection{Models for two Groups}
		\label{sec:models}


We assume in the rest of the paper that we have two groups of observations $G_1$ and~$G_2$\footnote{We often use $G_1, G_2$ instead of $P,N$ for the sake of unified notations ($n_i$ for the size of $G_i$, and such); it should be understood that $G_1 = P$ and $G_2=N$.} and that they are such that $G_1 \cap G_2 = \emptyset$.

The parameters describing the sizes of instances are as follows: 
$n = | G |$ is the total number of observations; 
$n_i = | G_i |$ is the number of observations in the group $G_i$; 
$k$ is the size of the projection $\pi_\Y (G)$ of the whole group on the subset $\Y$ of attributes; 
$k_i$ is similarly the size of the projection $\pi_\Y (G_i)$ of the group~$G_i$.

We shall consider different models, according to the available  information on the sizes of the groups and of their projections, and to whether the assumption $\pi_\Y (G_1) \cap \pi_\Y (G_2)= \emptyset$ holds.

\subsubsection{Which Sizes do we know?}
\label{sec:known-sizes}

For a given instance, we can assume that its number $n$ of observations is always known; of course $n \geq 1$.
When dealing with two subgroups of observations, in most cases we can assume that we known the sizes $n_1$ and $n_2$ of the positive and negative groups; again $n_1, \, n_2 \geq 1$ and $n \leq n_1 + n_2$, and the equality holds if the groups are disjoint.

Let us fix a candidate subset $\Y$ of attributes, either for the \MCP\ or for the \Patterns\ problem.
Depending on our data, we may know, or not, the sizes of the projection on~$\Y$, i.e., the number of distincts values, for the whole group of observations, and for the two subgroups.
We denote these sizes respectively by~$k$ and by $k_1, \, k_2$.
The size of the projection for the whole instance is such that $1\leq k \leq n$; $k_1$ and $k_2$ satisfy similar inequalities w.r.t. $n_1$ and $n_2$; finally $k \leq k_1 + k_2$.

Table~\ref{table:differents-cas} sums up the different cases that can formally be defined.
Not all of them will happen in real-life data but they matter for intermediate computations when computing actual probabilities (e.g., see Section~\ref{sec:two-groups-models-I}).

\begin{table}[h!]
\begin{center}
\begin{tabular}{|r||r|r|r|r|}
\hline
Case & $n$ & $n_1,\, n_2$ & $k$ & $k_1, \, k_2$	 \\ 
\hline 
{\bf (A)} & $\surd$ &  &  &   \\ 
{\bf (B)} & $\surd$ & $\surd$ &  &   \\ 
{\bf (C)} & $\surd$ &  & $\surd$ &   \\ 
{\bf (D)} & $\surd$ &  & $\surd$ & $\surd$  \\ 
{\bf (E)} & $\surd$ & $\surd$ & $\surd$ &   \\ 
{\bf (F)} & $\surd$ & $\surd$ & $\surd$ & $\surd$  \\ 
\hline 
\end{tabular} 
\end{center}
\caption{\label{table:differents-cas}
Summary of the known information on data sizes and on the sizes of projection on a given subset $\Y$ of attributes, from Case~{\bf (A)} where only the number of instances is known, to Case~{\bf (F)} where all the sizes are known.
}
\end{table}

\subsubsection{Is the Intersection of the Projections empty?}
\label{sec:empty-or-not}

Once an algorithm has proposed a partition of the set of attributes $\X$ into two disjoint subsets $\Y$ and $\Z$, we can consider, for example, the theoretical probability that the subset $\Y$ is a pattern. 
Therefore a basic assumption in such a case is that the two groups $G_1$ and $G_2$ and their projections $\pi_\Y (G_1)$ and $\pi_\Y (G_2)$ are disjoint.
This depends on the partition $\{ \Y, \Z \}$ of the set of attributes, hence we denote this model by M$1(\Y)$.

However, we can also try and guide an algorithm searching for a relevant partition of~$\X$. 
Such an algorithm may have to consider whether a given subset, or rather a given size of the subset~$\Y$, has a significant probability of allowing two groups to be disjoint. 
Here we need a model that allows for partitioning the data in two disjoint groups $G_1$ and $G_2$ whose projections are \emph{not} required to be disjoint: this is our second model M2($\Y$)\footnote{
The actual subset $\Y$ of Boolean attributes does not depend on the precise attributes but only on the number of such attributes: we might denote it by M2($| \Y |$), but keep the simpler notation M2($\Y$),which also emphasizes the parallelism with M1($\Y$).
}.

A third model M3 would consider data partitioned in two non-disjoint groups $G_1$ and $G_2$, with or without partition of the attribute set.
We only mention it for the sake of completeness, and because we can also compute probabilities in such a model, but do not explore it further in this paper.

\begin{figure}[h]
    \centering
    \includegraphics[width=4.5cm]{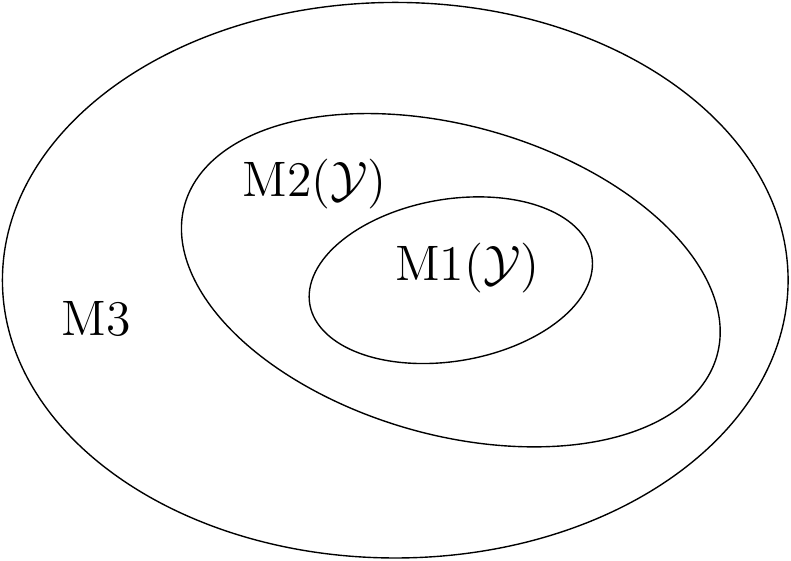}
    \caption{The three possible models for two groups $G_1$ and $G_2$: 
    M1($\Y$) and M2($\Y$) depend on a set $\Y$ of attributes and satisfiy  $G_1 \cap G_2 = \emptyset$;
    moreover $\pi_\Y (G_1) \cap \pi_\Y (G_2) = \emptyset$ in M1($\Y$); 
    finally there is no restriction on $G_1$ and $G_2$ in M3.}
    \label{fig:3modeles}
\end{figure}


	\subsection{How to Address Practical Questions}
      \label{sec:overview}


We now consider how combining the known information on sizes of Section~\ref{sec:known-sizes} and the eventual restrictions on intersections of Section~\ref{sec:empty-or-not} helps us to explore many probabilistic features of instances. 

	\subsubsection{Model M1($\Y$) for Algorithms Improvements and Analysis of Results}
	\label{sec:model_M1}

Several questions arise when dealing with real data. 
We recall that we restrict ourselves to two non-intersecting groups $P$ (positive) and $N$ (negative) and that here the intersection of the projections of~$P$ and~$N$ on the candidate subset~$\Y$ of attributes is also empty.

If a solution $\Y$, i.e., a subset of attributes, reduces the size of the projections of the groups on its attributes w.r.t. the initial group sizes, that means that the initial knowledge has been ``compiled'', in some sense, into a more compact formulation.
Of course this is related to learning, especially LAD, concepts.

\begin{enumerate}

\item {\bf Evaluating the reduction in the projection on a subset of attributes, for MCP.}
Given two groups $P$ and $N$ of sizes $n_1$ and $n_2$, what is the conditional probability $Pr(|\pi_\Y| = k /n_1, n_2, |\Y|)$  of having a projection of size $k$ on a set of attributes $\Y$, knowing the sizes $n_i$ of the groups and $|\Y|$ of the attribute subset~$\Y$? 
For given $n_1$ and $n_2$, we can plot $k$ as a function of the number of attributes; this can be used to evaluate the reduction effect of the projection, and is directly linked to the nature of the Boolean formula that can be extracted from $\Y$: smaller values of $k$ are expected to provide simpler formula. 
This reduction aspect is addressed in Section \ref{sec:calculs_difficult}.

\item {\bf Identification of difficult instances.}
Due to lack of information, in particular when not enough data are available, the initial structure of the instance may remain almost the same after the reduction process induced by $\Y$. 
We identify such instances as ``difficult''. 
The ideal case of reduction leads to a completely defined Boolean function: in such a case the projection on $\Y$ often induces an important change in the relative sizes of the initial groups. 
Difficult instances do not meet these conditions, but are not drawn from uniform random distributions. 
Computing the probability of a certain structure, in terms of the respective reductions of relative sizes for the initial groups under the assumption of a uniform distribution, helps us to point out these difficult instances. 
We examine this point in Section~\ref{sec:difficult}.

\item {\bf Finding a bound on the number of selected attributes, for algorithms solving MCP.}
A related question is to get a ``reasonable'' bound on the number of attributes $|\Y|$ for which we seek a  characterization.
In particular this allows us to get interesting bounds for Branch and Bound algorithms \cite{ChambonBLS18}; these bounds can then be used as starting point for an exhaustive search of solutions. 
The computation time may be large, and estimating a reasonable bound is particularly interesting.
This corresponds to checking which percentage of instances satisfy the partition of attributes into~$\Y$ and~$\Z$, and is handled in Section~\ref{sec:guess}.

\item {\bf Reliability, or robustness.}
Here we address the question of finding a set of attributes  $\Y$ such that all the possible values of $\Y$ appear: 
$| \pi_\Y | = 2^{|\Y|} = d_\Y$. 
If such a set~$\Y$ does exist, it means that the underlying partially defined Boolean function is indeed completely defined on the group~$P$ of positive observations.
Computing the probability that this happens provides information that may be used to establish a bound in order to search for solutions that contain more attributes than minimal solutions, but may be more reliable in some cases:
we expect the result to guide the search for reliable characterizations.
Section~\ref{sec:calculs_robust} deals with this question.

\item {\bf Patterns.}
Given two groups $P$ and $N$ of positive and negative observations, with respective sizes $n_1$ and $n_2$, what is the probability of a projection of size $k$ on a set $\Y$ of attributes, such that $k_1 = | \pi_\Y (G_1) |$ is as small as possible, preferably equal to~1 -- which happens when  the set of attributes $\Y$ is a pattern ? 
This corresponds to the probability that for a set $\Y$ of given size, the projection $\pi_\Y (G_1)$ has size $k_1$ with $k_1 =1$ when $\Y$ is a pattern, or $k_1$ has a  small value (2, 3, ..., extension of the pattern case), and is done in Section~\ref{sec:calculs_patterns}.

\end{enumerate}

The links between the problems we have just discussed and various probabilities are summed up in Table~\ref{table:questions-vs-coeffs}, which also expresses these probabilities in terms of suitable generating function coefficients (see Section~\ref{sec:two-groups-models-I} for the definitions, explicit expressions, and computations of these coefficients).

\begin{table}
\begin{center}
 
 \begin{tabular}{|c|c|}
\hline
Question & Probability   \\
\hline \hline
  Reduction by  $\Y$  & $Pr (k / n_1, n_2)$ 
  \\  \hline 
 Global structure of the instance after computation of $\Y$ 
  & $Pr (k_1,k_2 / n_1, n_2)$ 
  \\  \hline
  Reasonable bound for $|\Y|$ & $Pr(n_1,n_2 / n  ) $ 
  \\ \hline
  Reliability of size  $|\Y|$  & $Pr (k=n / n_1, n_2)$  
  \\ \hline
 Covering of $P$ by $r$ pattern & $Pr(k_1=r / n_1, n_2)$ 
  \\  \hline
 \end{tabular}
 
\end{center}

\caption{\label{table:questions-vs-coeffs} Summary of the questions for Model M1($\Y$), with $\Y$ a potential solution.
 }
\end{table}

	\subsubsection{Model M2($\Y$) for Improving Search Algorithms}

We now turn to the case where the two groups $P$ and $N$ (or $G_1$ and $G_2$) are disjoint, but their projections on some subset $\Y$ of attributes may intersect.

The main purposes here are to {\bf restrict the exploration space} when looking for a candidate subset~$\Y$ of attributes and to be able to consider {\bf data updates}.
For example, there maybe errors in a few observation that leads to the intersection of the projections being non-empty, but with very few elements; this possibility of such an error would correspond to a very low probability for these values of the intersection size under the already-chosen size~$|\Y|$. 
Or we may have found a suitable subset~$\Y$ for which the projections of the two subgroups do not intersect, but adding a new observation in one of the groups may cause the projections $\pi_\Y (P)$ and $\pi_\Y (N)$ to share a value.
On the other hand, a large size for the intersection of the projections would lead to questioning the partition of observations into the two groups $P$ and~$N$.

To address these purposes, we focus on the {\bf size of the intersection $\pi_\Y (P) \cap \pi_\Y (N)$}.
Assume that we obtain $|\Y|$ from an algorithm, or that we have an a-priori upper bound on $|\Y|$: we have to satisfy some condition on the number of attributes, without (yet) knowing the exact set of attributes. 

To this effect, we compute the probability $Prob( \pi_\Y (G_1) \cap \pi_\Y (G_2) = \emptyset / |\Y| = \ell )$ that the two groups have disjoint projections for a known number $\ell$ of simple attributes, which we gather together into the aggregated attribute~$\Y$. 
Choosing various values of~$\ell$ can help us to find the value(s) that maximize this probability.

A related question is to compute the probability that the intersection of the projections is not empty, or the probability that it is of known, small size (e.g., 1, 2, ..., 5).
This would also help in identifying the ``right'' size of~$\Y$, or at least a suitable range for it, and thus to speed up the search for the subset~$\Y$.
The probability than an update, by attributing observations to existing groups, creates projections that are no longer distinct is also related to the intersection size.
We consider how the mathematical results of former sections allow us to explore such questions in Section~\ref{sec:ModelM2-calculs}.

\section{Computational Examples}
\label{sec:compex}

The purpose of this section is to illustrate on real data instances the possible ways to use our mathematical models described in Section~\ref{sec:overview}. 
Our main focus is on Model M1, since it corresponds to practical situations where we are interested in computing solutions (i.e., non-intersecting projections of the groups on the computed solution $\Y$):
Section~\ref{sec:base-observations} describes the real instances on which we apply our models; 
then Section~\ref{sec:ModelM1-calculs} gives numerical results for Model~$M1$ with Section~\ref{sec:summary-MCP} summing what we can learn from these results.
Finally Section~\ref{sec:ModelM2-calculs} deals with Model~$M2$.

	\subsection{Data Instances}
	\label{sec:base-observations}
	

Our data instances come from previous case studies \cite{Chhel2013,ChambonBLS18,ChambonLSB19a}. 
Instances are sets of patho\-genic bacterial strains, observations are bacteria, and attributes represent genes (e.g., resistance gene or specific effectors, i.e., genes that we expect to be important in the characterization of the group of positive observations). 
These bacteria are responsible for serious plant diseases and their identification is thus important.
Groups are sets of bacteria that induce the same effects on the plants, so-called {\em pathovar}.
The main challenge is to characterize groups of bacteria using a limited number of genes, in order to design simple and cheap diagnosis tests \cite{Boureau2013}. 
In the original data several groups are considered; in our model this corresponds to having more than two groups~$G_i$.
We have considered the first group of bacteria as the positive group and the union of the other groups as the negative group; when considering any other group as the positive group, the results are similar to the ones presented below.

The characteristics of the biological instances are described in Table~\ref{tab:instances}, while Table \ref{tab:notations} recalls the main notations that will be used in the computations. 
Table~\ref{tab:instances} also gives the total number of attributes and the number of attributes in solutions found by the two algorithms used for solving the \MCP\ and \Patterns\ problems under the model M1($\Y$).
For \MCP, this is the number of attributes in minimal solutions; for \Patterns, this is the minimal number of attributes that must be used to fully cover the group $P$ of positive observations, using patterns. 
These last values are given in order to highlight that a full covering by a set of patterns may require more attributes that the minimal characterization. 
When the numbers of attributes in a minimal solution and in patterns are equal, it corresponds to the instances where $\Y$ is a unique pattern for $P$.
Note that, in both cases, several optimal solutions may exist for the same instance.

\begin{table}[h]
\begin{center}
\begin{tabular}{|l||c|c|c|c|c|}
\hline
Instance & \# Obs. & Positive & \# Attr. & \# Attr.  & \# Attr. \\
 &    & group size & (total) & (sol. MCP) & (sol. Patterns) \\
\hline
\emph{rch8} & 132 & 5 & 37  & 3  & 3\\
\hline
\emph{ra\_rep1} & 112 & 38 & 155& 12 & 22\\
\hline
\emph{ra\_rep2} & 112 & 37 & 73 & 11 & 19\\
\hline
\emph{ralsto} & 73 & 27 & 23& 5 & 5\\
\hline
\emph{ra100\_phv} & 101 & 21 & 50& 2  & 2\\
\hline
\emph{ra100\_phy} & 105 & 31 & 51 & 3  & 4\\
\hline
\emph{ra\_phv} & 108 & 22 & 70 & 2 & 2\\
\hline
\emph{ra\_phy} & 112 & 31 & 73 & 3 & 4\\
\hline
\end{tabular}
\end{center}
\caption{\label{tab:instances} Main characteristics of the eight instances: initial data, and solutions found by \MCP\ and \Patterns.}
\end{table}

\begin{table}[h]
    \centering
    \begin{tabular}{|c c|}
    \hline
        $n$ &  total number of observations\\
        $n_i$ & number of observations in group $G_i$, $i=1,\, 2$\\
        $|\X|$ & total number of attributes\\
        $|\Y|$ & number of attributes for the optimal solutions\\
        $nb_{sol}$ & number of optimal solutions (non dominated)\\
        $k$ & size of the projection  $\pi_\Y(G) $ on the whole instance\\
         &(average value if multiple solutions)\\
        $k_i$ & size of the projection  $\pi_\Y(G_i) $ on group $G_i$, $i=1,\, 2$\\
    \hline
    \end{tabular}
    \caption{Reference table for notations}
    \label{tab:notations}
\end{table}


\subsection{Computations using Model M1($\Y$) (disjoint projections)}
	\label{sec:ModelM1-calculs}
	

In this section we compare the actual results of Table~\ref{tab:instances}, obtained by the algorithms we used to solve the problems \MCP\ and \Patterns, to the theoretical results predicted by our modelization for M1($\Y$).
In this model the candidate subset of attributes $\Y$ is known, and the two groups $G_1$ and $G_2$ are disjoint as well as their projections $\pi_\Y (G_1)$ and $\pi_\Y (G_2)$.
We also know the sizes $n_1$ and $n_2$ of the two groups and the total size $k$ of the projection: we are in case~{\bf (E)} of Table~\ref{table:differents-cas}.
If furthermore we know the sizes of the group projections, the relevant case of that table is now~{\bf (F)}.

Table~\ref{table:realdata} presents in further detail the results obtained by the \MCP\ and solving algorithm.
When there is more than one solution ($nb_{sol} > 1$), the numbers $k$ and $k_1$ are the average values of the projection sizes for these solution, and are not integer-valued.
\begin{table}[h!]
\begin{center}
\begin{tabular}{|l||r|r|r|r|r|r|r|r|r|}
\hline 
Instance & $n$ & $n_1$ & $n_2$ & $|\X|$ & $|\Y|$ & $nb_{sol}$ & $k$ & $k_1$ \\ 
\hline 
\emph{rch8} & 132 & 5 & 127 & 37 & 3 & 1	&8	&1	\\ 
\emph{ra\_rep1} & 112 & 38 & 74 & 155 & 12 & 134	&95,30	&30,17\\ 
\emph{ra\_rep2} & 112 & 37 & 75 & 73 & 11 & 106	&89,5	&28,35	\\ 
\emph{ralsto} & 73 & 27 & 46 & 22 & 5 &5	&17	&7,8\\ 
\emph{ra100\_phv} & 101 & 21 & 80 & 50 & 2 & 1	&4	&1\\ 
\emph{ra100\_phy} & 105 & 31 & 74 & 51 & 3 & 1	&8	&5\\ 
\emph{ra\_phv} & 108 & 22 & 86 & 70 & 2 & 1	&3	&1\\  
\emph{ra\_phy} & 112 & 31 & 81 & 73 & 3 &1	&8	&5\\ 
\hline 
\end{tabular} 
\end{center}
\caption{\label{table:realdata}
Results for two groups, as computed by the MCP algorithm: $\Y$ is the number of attributes of optimal solution(s), $nb_{sol}$ the number of solutions with $|\Y|$ attributes, $k$ and $k_1$ the respective sizes of the projections on~$\Y$ of the whole observation and of the set of positive observations.}
\end{table}

	\subsubsection{Structural Properties of the Instances}
	\label{sec:calculs_difficult}

Under our probabilistic assumptions (see Section~\ref{sec:hypotheses-probabilistes}), the conditional probability that the projection of the instance on~$\Y$ has size~$k$, knowing that its two groups have sizes~$n_1$ and~$n_2$, is obtained by dividing the number of instances with two groups of respective sizes $n_1$ and $n_2$ and projection size~$k$, by the total number of instances with groups of sizes $n_1$ and~$n_2$.
Expressing this probability in mathematical terms is done in Section~\ref{sec:two-groups-models-I}; with the notations of that section this is
\[
Pr (k / n_1, n_2) = \frac{\gamma_{k;n_1,n_2}}{\rho_{n_1,n_2}}.
\]
Now set $k_{max} = \inf(n,d_\Y)$ for the values of Table~\ref{table:realdata}; this is the maximal possible value of the projection size.
For all instances except \emph{ra\_rep1} and \emph{ra\_rep2}, $k_{max} = d_\Y$; for the two instances \emph{ra\_rep1} and \emph{ra\_rep2}, $k_{max} = n$. 

To compute the probability that the projection on the subset $\Y$ of attributes is of size~$k$ for the eight instances of our example database, we use the value of~$k$ already obtained by the solving algorithms; these values are the ones given in Table~\ref{table:realdata}.
When we have found more than one solution, we take for $k$ the average value of the actual projection sizes for the different solutions.
This is not always an integer and we have rounded it to~95 for \emph{ra\_rep1} and to~89 for \emph{ra\_rep2}.

The probability that an instance has $k$ different values on~$\Y$, i.e., that we cannot reduce the data through projection, is given in Table~\ref{table:proportional}. 
Results very close to~$1$ are written as a difference $1-\epsilon$, with a very small value~$\epsilon$.

\begin{table}[htbp]
\begin{center}
\begin{tabular}{|l||r|r|r|r|l|l|l|l|l|}
\hline 
Instance & $n_1$ & $n_2$ & $d_\Y$ & $d_\Z$ &$k_{max}$ & $k$ & Pr($k_{max}$) & Pr($k$) \\ 
\hline 
\emph{rch8} &  5 & 127 & 8 & $2^{34}$ & 8 & 8 & 1 - $2.2 \, 10^{-8}$& same value \\ 
\emph{ra\_rep1}  & 38 & 74 & 4096 & $2^{143}$ & 112 & 95.1 & 0.429 & 1.981 $10^{-18}$ \\ 
\emph{ra\_rep2}  & 37 & 75 & 2048 & $2^{62}$ & 112 & 89.49 & 0.176 &2.792 $10^{-21}$ \\ 
\emph{ralsto}  & 27 & 46 & 32 & $2^{17}$ & 32 & 17 &0.115 & 7.318 $10^{-16}$ \\ 
\emph{ra100\_phv}  & 21 & 80 & 4 & $2^{48}$ & 4 & 4& 1 - $ 2.5 \, 10^{-14}$  & same value\\ 
\emph{ra100\_phy}  & 31 & 74 & 8 & $2^{48}$ & 8 & 8 & 1  - $5 \, 10^{-6}$ & same value \\ 
\emph{ra\_phv}  & 22 & 86 & 4 & $2^{68}$ & 4 & 3 & 1 - $2.2 \, 10^{-15}$  & 2.154 $10^{-15}$ \\ 
\emph{ra\_phy}  & 31 & 81 & 8 & $2^{70}$ & 8 & 8 & 1  - $2 \, 10^{-6}$ & same value \\ 
\hline 
\end{tabular} 
\caption{\label{table:proportional}Results for the probability that the projection on $\Y$ has size~$k$. 
}
\end{center}
\end{table}

The theoretical probability $Pr(k_{max})$ of having a projection of maximal size is very close to $1$ in the first case and the four last cases, and smaller but not close to~0 in the remaining three cases. 
Note also that for all these instances, there is only one solution. 

For \emph{ra\_phv,} the probability goes from very close to~1, to very close to~0, when we substitute $k_{real} = 3$ to the theoretical value $k_{max} = d_\Y$; this strongly suggests that our assumption of independence and uniformity on attributes does not hold. 
Therefore, this instance has certainly some specific inner structures and the solution found by the algorithm is especially interesting. 

Consider now the three cases where \MCP\ gives several solutions.
For \emph{ra\_rep1} and \emph{ra\-rep2,} the probability that $k_{max} = d_\Y$ is null (there are not enough observations to cover all possible values of~$\Y$); when we consider the probability that $k_{max} = n$, i.e., that all the $\Y$~values in the instance are different, we obtain a significant, non-null probability; changing $k_{max}$ to $k_{real}$ gives a probability that is now very close to~0. 

The same phenomenon happens for \emph{ralsto} when changing $k_{max}$ from 32 (i.e., $d_\Y$) to $k_{real}=17$. 
Notice that the reduction from $k_{max}$ to $k_{real}$ constitutes thus an important reduction and really makes sense.
The resulting projections do not coincide with what we would expect from a uniform distribution.


\medskip
It is also possible to study the variation of the probability that the projection on $\Y$ has size~$k$ according to that size.
We give an example of such a study for the instance \emph{ra\_rep1}.
Numerical results show that the probability is negligible at first: null for $k=1$ and equal to $8.679 \, 10^{-398}$ for $k=2$, then growing slowly until it becomes significant; for example the probability that $k=100$ is still only  $2.279 \, 10^{-11}$; only the last few values for $k$ have a significant probability of appearing.
See Table~\ref{table:ra_rep1_probas}.

\begin{table}[h!]
\begin{center}
\begin{tabular}{|l|r||l|r|}
\hline 
$k$ & Pr( $k/n_1,n_2$ ) & $k$ & Pr( $k/n_1,n_2$ )\\
\hline
90 & $1.873 \; 10^{-26}$ & 102 & $7.325 \; 10^{-9}$ \\
91 & $8.866 \; 10^{-25}$ & 103 & $1.095 \; 10^{-7}$ \\
92 & $3.878 \; 10^{-23}$ & 104 & $1.432 \; 10^{-6}$ \\
93 & $1.565 \; 10^{-21}$ & 105 & $1.619 \; 10^{-5}$ \\
94 & $5.811 \; 10^{-20}$ & 106 & 0.000156 \\
95 & $1.982 \; 10^{-18}$ & 107 & 0.00125 \\
96 & $6.189 \; 10^{-17}$ & 108 & 0.00814 \\
97 & $1.765 \; 10^{-15}$ & 109 & 0.0412 \\
98 & $4.577 \; 10^{-14}$ & 110 & 0.153 \\
99 & $1.076 \; 10^{-12}$ & 111 & 0.367 \\
100 & $2.279 \; 10^{-11}$ & 112 & 0.429 \\ 
101 & $4.329 \; 10^{-10}$ & & \\
\hline
\end{tabular}
\caption{\label{table:ra_rep1_probas}Probability that the projection on $\Y$ has size~$k$, for \emph{ra\_rep1} and with a size of projection $k=90...112$.}
\end{center}
\end{table}

\subsubsection{Difficult Instances}
\label{sec:difficult}

Assume that $\Y$ has been chosen; we may be interested in comparing the initial size of the positive group ($n_1$) with the size of its projection on $\Y$. 
For the seven instances of Table~\ref{tab:instances}, Table~\ref{table:ratios} presents the percentages of positive observations for the whole observations (ratio $n_1 / n$) and for their projection on~$\Y$ (ratio $k_1 / k$). 
For the instances \emph{ra\_rep1}, \emph{ra\_rep2} and \emph{ralsto}, there is more than one candidate $\Y$ set of attributes, which means that the values given in Figure~\ref{table:ratios} are computed from the average values of $k$ and $k_1$ obtained for those different sets~$\Y$.

\begin{table}[h!]
\begin{center}
\begin{tabular}{|l||l|l|}
\hline
Instance &  $n_1/n$  & $k_1/k$\\ 
\hline 
{\em rch8} 	&	0.0379 	&	0.125 	\\
{\em ra\_rep1} 	&	0.3393 	&	0.3125 	\\
{\em ra\_rep2} 	&	0.3303 	&	0.3170 	\\
{\em ralsto} 	&	0.2079 	&	0.4588 	\\
{\em ra100\_phv}  	&	0.2952 	&	0.25 	\\
{\em ra100\_phy} 	&	0.2037 	&	0.625 	\\
{\em ra\_phv}  	&	0.2037 	&	0.3333 	\\
{\em ra\_phy}  	&	0.2768 	&	0.625 	\\
\hline 
\end{tabular} 
\end{center}
\caption{\label{table:ratios}
Comparison of the percentage of observations in $G_1$ (ratio ${n_1}/{n}$) and in $\pi_\Y(G_1)$ (ratio ${k_1}/{k}$) for the minimal attribute set $\Y$ computed by the MCP algorithm.
}
\end{table}

When $|\Y|$ is small, the structure of the instance changes after projection  (ratios ${n_1}/{n}$ vs $k_1/k$) , since in these cases the resulting projected table is very compact and the importance of the characterization of positive observation increases (this is particularly important in instance {\em rch8}). 
For more difficult instance (in terms of reduction), we observe that the ratio is almost the same (instances {\em $ra\_rep$}). 
In these cases, the amount of observations in not sufficient to extract more compact representation and finally the reduction process has a linear effect of the data. 
Note that the probability of having a projection of size $k$ (see Table \ref{table:proportional}) is then very low. 
This suggests that the computed probability should alert us on the difficulty of the instance to check these reduction ratios.
Instances {\em ralsto} and {\em ra\_phv} fall in-between these two cases.  

\medskip
Concerning the difficulty of reduction of the instances: if the number of observations is too small as regards the number of required attributes, then the behaviour of the instance is far from the uniform model and the characterization we obtain can be assessed as a relevant process. 

Concerning instances for which $|\Y|$ is small: due to the small size of the projections, the uniform model meets the real cases since we often reach completely defined functions. 
In these cases, a uniformly distributed instance, restricted to $\Y$,  with enough observations will lead to such completely defined function due to the small value of $d_\Y$. 
Otherwise, given a uniform distribution, the reduction is unlikely to occur since we have no knowledge available to reduce the representation of the observation, while on our instances the method provides some useful reduction on well chosen attributes. 
Our model can be used to study such properties on instances. 
More precisely, we are interested in being able to identify instances where the data do not contain any {\em learnable} characterization due to the fact that  there are either issued from a normal distribution (e.g., bad choice of attributes) or that the number of observations is too small to detect relevant patterns of combination of attributes.

	\subsubsection{Finding a Bound for $|\Y|$}
	\label{sec:guess}

Computing the proportion of instances with two groups of sizes $n_1$ and $n_2$ among those of global size~$n=n_1+n_2$  may give indication on the ``best'' size for~$\Y$ (this is the ratio $\rho_{n_1,n_2} / \rho_n$ of the numbers defined in Section~\ref{sec:two-groups-models-I}, Cases~B and~A).
Therefore, it gives us an indication whether the chosen values corresponds to a reasonably likely situation for having disjoint  groups.    

\begin{figure}
\centering
\includegraphics[width=6cm]{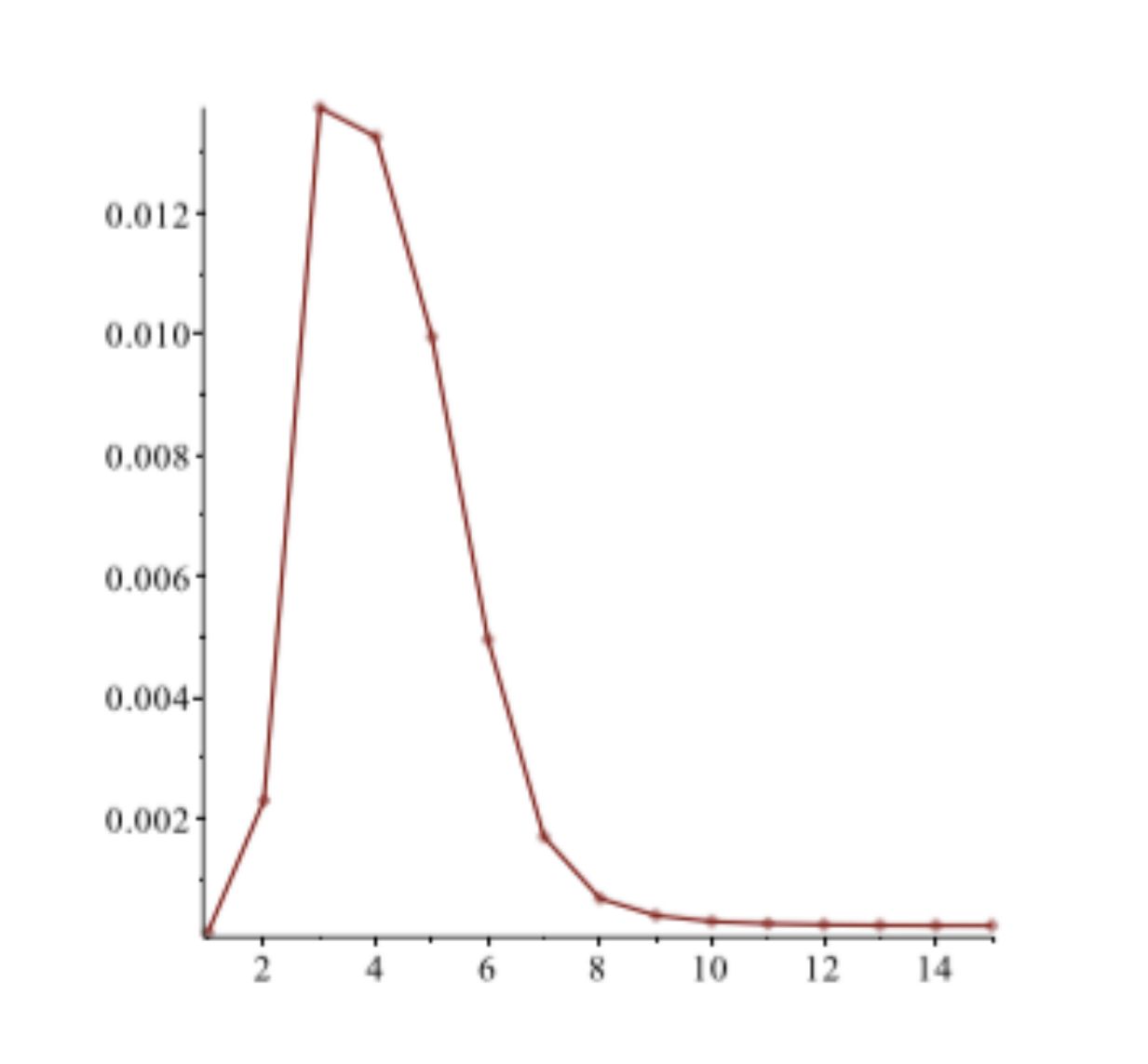}
\\
\caption{Plot of the ratio ${\rho_{n_1,n_2}}/{\rho_n}$ against $|\Y|$ for \emph{ra\_rep1}.
The optimal number of attributes found by \MCP\ for this instance is~12.}
\label{fig:rarep1_curve}
\end{figure}

\medskip
The numerical computations show that the ratio ${\rho_{n_1,n_2}}/{\rho_n}$ is unimodal when $|\Y|$ increases: it increases; then one or two values giving the maximal ratio; then it decreases; see Figure~\ref{fig:rarep1_curve} for an example\footnote{An interesting question would be to consider whether unimodularity always holds and prove it mathematicallly; we have refrained from considering it, so as to keep the present paper within reasonable size.}. 

Table \ref{tab:ratio_rob_rch8}) gives this ratio for the instance \emph{rch8} around the actual value $|\Y| = 3$. 
The maximal ratio is obtained for $\Y$ with four attributes, and is only twice the ratio for three attributes, but the values for a smaller or larger number of attributes fall quickly.
\begin{table}[h]
\begin{center}
\begin{tabular}{|r|r|}
\hline 
$|\Y|$& Ratio $\rho_{n_1,n_2} / {\rho_n}$
\\ \hline
2 & $1.025 \; 10^{-11}$  \\
3 & $1.27 \; 10^{-5}$  \\
4 & $2.06 \; 10^{-5}$  \\
5 & $2.15 \; 10^{-8}$  \\
\hline
\end{tabular}
\end{center}
\caption{\label{tab:ratio_rob_rch8} Values of the ratio ${\rho_{n_1,n_2}}/{\rho_n}$ for \emph{rch8}.}
\end{table}

In all but two of the instances (\emph{ra\_rep1} and \emph{ra\_rep1}), the optimal number of attributes we found with this approach either is the one given in Table~\ref{tab:instances} or differs by~1 from this number. 
Nevertheless this is not the case for all instances, e.g., for \emph{ra\_rep1} the situation is different: the actual value $|\Y| = 12$ is very far from the one giving the maximal ratio, which occurs for $|\Y| = 3$ (see again Figure~\ref{fig:rarep1_curve}).

Of course, the properties we commented on are related to the sizes $| \Y |$ and $n,n_1,n_2$. 
If $|\Y|$ is too small and the values $n,n_1,n2$ large enough, then it becomes impossible to obtain disjoint groups after projection and Model $M_1(\Y)$ does not hold. 
Moreover, if we select a theoretical value $\Y$ far from the optimal solution, the same problem occurs since $n$, $n_1$ and $n_2$ are fixed. 

When the optimal number of attributes $|\Y|$ is small w.r.t. the other parameters of the instances (i.e. instances where the reduction is important), the computations agree with the results obtained from solutions. 
That does not mean that our uniform model always holds, but that it can still provide a useful bound on~$\Y|$.

	\subsubsection{Robustness}
	\label{sec:calculs_robust}

We consider that $\Y$ is robust if the projection $\pi_\Y(G)$ has size $k = d_\Y$ (this is Case~{\bf (C)} of Table~\ref{table:differents-cas}).
In this case, the partially defined Boolean function turns into a fully defined Boolean function. 
Therefore, from a learning point of view the proposed characterization is able to handle all possible incoming data and assign it to a group.

\paragraph{Without groups}

For our first computations, we do not consider the partition of data instances in two groups and simply compute the probability that all the possible values of the domain $d_\Y$ appear in the instance and in the projection; again using the notations of Section~\ref{sec:two-groups-models-I}, this is
\[
Pr (| \pi_\Y (G) | = d_\Y / n) = \frac{\beta_{d_\Y;n}}{\rho_n}.
\]
If $d_\Y > n$, robustness is simply impossible: there are not enough observations to cover all the possible values of~$\Y$; if $n >> d_\Y$, robustness is highly likely; and the only case in which we expect robustness to have a probability distinct from both~0 end~1 is when $n$ and $d_\Y$ are roughly of the same order.
The numerical results given in the next-to-last column of Table~\ref{table:robustness} confirm this intuition.

\begin{table}[h!]
\begin{center}
\begin{tabular}{|l||r|r|l|r|r|}
\hline
Instance & $n$ & $|d_\Y|$ & $|d_\Z|$ & Pr(robust.) & Pr( robust., 2 groups)	 \\ 
\hline 
\emph{rch8} & 132 & 8 & $2^{34}$ & {1} - $1.7 \, 10^{-7}$ & {1} - $ 10^{-8}$ \\ 
\emph{ra\_rep1} & 112 & 4096 & $2^{143}$ & 0 & 0\\ 
\emph{ra\_rep2} & 112 & 2048 & $2^{62}$ & 0 & 0 \\ 
\emph{ralsto} & 73 & 32 & $2^{17}$ & 0.0209 &0.440\\ 
\emph{ra100\_phv} & 101 & 4 & $2^{48}$ & {1} $ - 10^{-12}$  & 1 - $5 \, 10^{-18}$\\ 
\emph{ra100\_phy} & 105 & 8 & $2^{48}$ &  {1} - $7 \, 10^{-6}$ & {1} - $3 \, 10^{-8}$ \\ 
\emph{ra\_phv} & 108 & 4 & $2^{68}$ & {1} $ -  \, 10^{-12}$  & {1} - $3 \, 10^{-19}$ \\ 
\emph{ra\_phy} & 112 & 8 & $2^{70}$ & {1} - $3\, 10^{-6}$ & {1} - $9 \, 10^{-9}$ \\ 
\hline 
\end{tabular} 
\end{center}
\caption{\label{table:robustness}
Robustness: probability that the projection on $\Y$ has size~$d_\Y$ (i.e., all the values of $D_\Y$ appear in the instance), without and with partition of the instance in two groups.
}
\end{table}

In two cases (\emph{ra\_rep1} and \emph{ra\_rep2}) the size $d_\Y$ (respectively $2^{11}$ and $2^{12}$) is much larger than~$n$, and it is impossible to have $k=d_\Y$. 
For the instance \emph{ralsto}, where $n$ and $d_\Y$ are roughly of the same order ($n=73$ and $d_\Y = 32$) and here the probability of robustness is small, roughly equal to 2.1\%; see the next-to-last column of Table~\ref{table:robustness}.
In the remaining cases, the probability of robustness is very close to~1; this is not a  surprise as $n$ is much larger than~$d_\Y$.

\paragraph{With two groups}

Here we know the group sizes $n_1$ and $n_2$, and the size $k=d_\Y$ of the projection. 
The probability of robustness is here
\[
Pr (| \pi_\Y (G) | = d_\Y / n_1, n_2) = \frac{\gamma_{d_\Y;n_1, n_2}}{\rho_{n_1,n_2}}.
\]
Again the only case for which the probability of robustness is not 0 or close to~1 occurs with \emph{ralsto}. 
In the other cases, the probability is either~0 or close to~1, and the existence of the two groups does not make a difference.

	\subsubsection{Patterns}
	\label{sec:calculs_patterns}

We consider here both cases: whether a single pattern or a given number of patterns cover one group.
This falls into case~{\bf (F)} of Table~\ref{table:differents-cas} and the mathematical expression of the conditional probability is
\[
Pr ( k_1 , k_2 / n_1, n_2) = \frac{\delta_{k_1, k_2;n_1, n_2}}{\rho_{n_1,n_2}}.
\]

    \paragraph{Single pattern}

There is a single value for $\Y$ common to all the observations of the positive group, i.e., $k_1=1$ (and of course $k_2 = k-1$); to obtain the probability that $\Y$ is a pattern, we fix $k_1=1$ and forget about $k_2$.

The results are given in Table~\ref{table:patterns-different-k}, column $\Pr(k_1 =1)$.
The probabilities are very close to either 1 in three cases (\emph{rch8,} \emph{ra100\_phv} and \emph{ra\_phv}) or~0 (all the other cases).
When comparing the theoretical results of Table~\ref{table:patterns-different-k} with those obtained by the \Patterns\ algorithm (see the last column of Table~\ref{tab:instances}), we observe that the high probability that $\Y$ is a pattern occurs in cases where this condition is satisfied in real data (i.e. $k_1=1$). 
Therefore, this computation is particularly interesting to help in focusing on relevant sizes when searching for~$\Y$.

\begin{table}[htbp]
\begin{center}
\begin{tabular}{|r||r|r|l|l|l|l|l|}
\hline 
Instance &  $n_1$ & $n_2$ & $d_\Y$ & $d_\Z$ & Pr($k_1 =1$) & Pr($k_1 =2$)  & Pr($k_1 =3$)  \\ 
\hline 
\emph{rch8} & 5 & 127 & 8  & $2^{34}$ & 1- $10^{-6}$ & $3.303 \; 10^{-7}$ & $2.903 \; 10^{-16}$ \\ 
\emph{ra\_rep1} &  38 & 74 & 4096 & $2^{143}$ &  $4.295 \; 10^{-134}$ & $2.374 \; 10^{-119}$ & $1.564 \; 10^{-109}$\\  
\emph{ra\_rep2} &  37 & 75 & 2048 & $2^{62}$  & $2.316 \; 10^{-119}$ & $3.140 \;10^{-105}$ & $6.764 \; 10^{-96}$\\ 
\emph{ralsto} &  27 & 46 & 32 & $2^{18}$  & $6.765 \; 10^{-28}$ & $3.118 \; 10^{-19}$ & $3.726 \; 10^{-14}$ \\ 
\emph{ra100\_phv} &  21 & 80 & 4 & $2^{48}$  & 1- $10^{-6}$ & $2.573 \; 10^{-8}$  & $7.073 \; 10^{-29}$ \\ 
\emph{ra100\_phy} &  31 & 74 & 8 & $2^{48}$ & $6.666 \; 10^{-6}$ & 0.557  & 0.443  \\  
\emph{ra\_phv} & 22 & 86 & 4 & $2^{68}$  & 1- $10^{-6}$ &
$4.517 \; 10^{-9}$  & $2.911 \; 10^{-31} $\\ 
\emph{ra\_phy} & 31 & 81 & 8 & $2^{70}$ & $2.882 \; 10^{-5}$ & 0.818 & 0.182 \\ 
\hline 
\end{tabular} 
\caption{\label{table:patterns-different-k}
Probability that $\pi_\Y (G_1)$ has size $k_1$.
The column for $k_1=1$ gives the probability that $\Y$ is a pattern.}
\end{center}
\end{table}

    \paragraph{Multiple patterns}

We now consider small values of $k_1$ to obtain the probability that $\pi_\Y (G_1)$ has size~$k_1$, according to the values of~$k_1$ from 1 to $n_1$.
This means that a single set of attributes~$\Y$ determines two or more patterns on \emph{the same attributes}. 

The results are presented in Table~\ref{table:patterns-different-k}, for $k_1$ from~1 (single pattern) to~3.
We observe that on \emph{ra100\_phv} and \emph{ra\_phv} the probability decreases quickly, and that we have the reverse phenomenon on \emph{ra100\_phy} and \emph{ra\_phy} for which a size $k_1$ equal to~2 or~3 is most likely.
These observations are coherent with real data.

\subsection{Summary of what we learned on the instances}
\label{sec:summary-MCP}

We consider in this section what we can learn for each of our example instances fromt he numerical data of Section~\ref{sec:ModelM1-calculs}.
A more general overview of the conclusions we may draw from such analyses is proposed in Section \ref{sec:scope}.

\subsubsection{Instance \emph{rch8} }
\label{instance:rch8}
This instance has a very small value for the size $n_1$ of the group~$G_1$, compared with the total number $n$ of observations.
Therefore $\Y$ could be expected to contain a large number of attributes to distinguish the positive observations from negative ones. 
Nevertheless, this is not the case: this  means that the positive observations share a strong common structure, represented by a unique pattern. 
The ratio ${k_1}/{k}$ increases compared to to ${n_1}/{n}$, leading to a compact table that corresponds to a completely defined Boolean function with one line that fully defines the positive group.  
This is somehow an ideal case for practitioners: full identification of bacteria can be achieved using $3$ attributes out of $37$.

The computation of the minimal set $\Y$ to reach the suitable optimal bound is costly.
Table \ref{tab:ratio_rob_rch8} gives the computed ratio $\rho_{n_1,n_2} / \rho_n$ for the actual value $|\Y| = 3$ and for a few values around it.
This ratio is maximal for $|\Y| = 4$; it is only twice the ratio for three attributes; but the values for a smaller or larger number of attributes fall quickly. 
This probabilistic estimation can be used to focus on a suitable search interval for $|\Y|$.

\subsubsection{Instances \emph{ra\_rep1} and \emph{ra\_rep2}}
\label{instance:rarep12}

Structural properties and difficult to observe:
for \emph{ra\_rep1} and \emph{ra\-rep2,} the probability that $k_{max} = d_\Y$ is null since there are not enough observations to cover all possible combinations of values generated by $\Y$ and it is not possible to reach a completely defined Boolean function here. 
Nevertheless, if we consider $k_{max} = n$ (i.e., all different projections of the initial observations on $\Y$ are different), we obtain a significant, non-null probability. 
This means that, under the uniform distribution assumption, we should reasonably expect to get no reduction after projection: the number of initial observations is very low compared to the number of attributes. 
Nevertheless, changing $k_{max}$ to $k_{real}$ gives a probability that is now very close to~0;  therefore this probabilistic estimation allows us to argue that the data set does not follow a uniform distribution. 

Moreover, for \emph{ra\_rep1} let us consider the variation of the probability that the projection on $\Y$ has size~$k$ for varying~$k$.
The numerical results of Table~\ref{table:ra_rep1_probas} show that the probability is negligible at first: null for $k=1$ and equal to $8.68 \, 10^{-398}$ for $k=2$, then growing slowly until it becomes significant; for example the probability that $k=100$ is still only  $2.28 \, 10^{-11}$; only the last few values for $k$ have a significant probability of appearing. 
This confirms the conclusion we stated above, that the instance does not correspond to values generated according to a uniform distribution.  

But we observe that the ratios $n_1/n$ and $k_1/k$ are almost the same (instances {\em $ra\_rep$}): the amount of observations in not sufficient to extract a more compact representation, and the reduction process has a linear effect of the data.

Bound for $|\Y|$: The optimal number of attributes given by the computation of $\rho_{n_1,n_2} / \rho_n$ is far smaller for other instances than for the instances \emph{ra\_rep1} and \emph{ra\_rep2} (cf. Figure~\ref{fig:rarep1_curve}). 
Here the situation is different: the actual value $|\Y| = 12$ for \emph{ra\_rep1} is very far from the one giving the maximal ratio, which occurs for $|\Y| = 3$. We do not detail the  \emph{ra\_rep2} case which is similar. Again, these instances have specific properties concerning their sizes and groups (see Table \ref{tab:instances}). 


\subsubsection{Instance \emph{ralsto}}
\label{instance:ralsto}
Structural properties:
a phenomenon similar to that for \emph{ra\_rep1} and \emph{ra\-rep2} happens, when changing $k_{max}$ from 32 (i.e., $d_\Y$) to $k_{real}=17$: there is a significant, non-null probability that all the $\Y$~values in the instance are different ($k_{max} = n$); changing $k_{max}$ to $k_{real}$ gives a probability that is now very close to~0. 
Notice that the reduction from $k_{max}$ to $k_{real}$ thus constitutes an important reduction and really makes sense.
The resulting projections do not coincide with what we would expect from a uniform distribution.  

Robustness:
For \emph{ralsto}, where $n$ and $d_\Y$ are roughly of the same order ($n=73$ and $d_\Y = 32$), the probability of robustness is small, roughly equal to 2.1\%; see the next-to-last column of Table~\ref{table:robustness}.

\subsubsection{Instances \emph{ra100\_phv} and \emph{ra\_phv}}
\label{instance:ra100phv}

Patterns: 
on Table \ref{table:patterns-different-k}, we observe that for \emph{ra100\_phv} and \emph{ra\_phv} the probability of having a pattern of size $1$ is very high. These estimation are fully coherent with real data, since an optimal set $\Y$ leads to a pattern of size $1$ that covers the whole set $P$.

Structural properties : For \emph{ra\_phv,} the probability of having a reduction of the initial set $P$ goes from very close to~1, to very close to~0, when we substitute $k_{real} = 3$ to the theoretical value $k_{max} = d_\Y$. 
Therefore the resulting reduction of the initial instance to only $3$ observations and a unique pattern confirms that this instance certainly has some specific inner structure, and the solution found by the algorithm is especially interesting to investigate by practitioners.

\subsubsection{Instances \emph{ra100\_phy} and \emph{ra\_phy} }
\label{instance:raphv}

Patterns: 
on \emph{ra100\_phy} and \emph{ra\_phy} the probability for having a single pattern on Table \ref{table:patterns-different-k} is very low but it increases as $k$ increases too. 
These observations are coherent with real data (see Table\ref{table:intro-resultat}, $5$ patterns are indeed required to cover $P$ if considering attributes in $\Y$). 
Therefore, the probabilistic estimation is pertinent here.


	\subsection{Computations using Model M2($\Y$)}
	\label{sec:ModelM2-calculs}
	

Here we no longer assume that the projections of the two groups are disjoint, although the groups still are.
We are interested in the probability that the intersection $ \pi_\Y (G_1) \cap \pi_\Y (G_2)$ is empty, or of small known  size, or possibly of bounded (small) size.
This is Case~{\bf (B)} of Table~\ref{table:differents-cas}.
The relevant formulae can be found in Section~\ref{sec:M2-size-intersection} and the numerical values of the probabilities are given in Table~\ref{tab:empty_int}.

\begin{center}
\begin{table}[h]
\centering
    \begin{tabular}{|l||l|l|l|l|}
    \hline
    Instance & Pr($r=0$) &  P ($r=1$) & Pr($r=2$) & Pr($r \leq  4$)\\
    \hline
      \emph{rch8}   & 1.05 $10^{-11}$ & 0.000244 & 0.0256 & 0.795 \\
       \emph{ra\_rep1}   &  0.50 & 0.350 & 0.117 & 0.999 \\
       \emph{ra\_rep2}   &  0.26  & 0.36 & 0.24 &  0.991 \\
       \emph{ralsto} & 2.5 $10^{-13}$ & 2.68 $10^{-11}$ & 1.24 $10^{-9}$ & 6.2 $10^{-7}$ \\
       \emph{ra100\_phv} & 9.2 $10^{-23}$ & 9.1 $10^{-13}$ & 2.86 $10^{-6}$ & 1 \\
       \emph{ra100\_phy} & 6.2 $10^{-27}$  & 6.67 $10^{-21}$ & 6.53 $10^{-16}$ &3.93 $10^{-8}$  \\
       \emph{ra\_phv} & 4.1 $10^{-24}$  & 2.27 $10^{-13}$ & 1.43 $10^{-6}$ &  1 \\
       \emph{ra\_phy} & 5.6 $10^{-28}$  & 1.04 $10^{-21}$ & 2.31 $10^{-16}$& 3.52 $10^{-8}$\\
         \hline 
    \end{tabular}
    \caption{Model M2($\Y$): Probability that the intersection $\pi_\Y (G_1) \cap \pi_\Y (G_2)$ is empty ($r=0$) or that its size $r$ has a small value.}
    \label{tab:empty_int}
    \end{table}
\end{center}

These results allow us to assess that most of instances are drawn from a data distribution that is far from uniform. 
Actually, for most instances the probability that the groups are disjoint is very low, while the different data sets have disjoint $P$ and $N$.
This confirms that the initial data was not drawn from uniformly independent distributed attributes.

When the data sets have few observations compared with their number of attributes, the uniformity  property cannot be verified.
This is the case for the two instances \emph{ra\_rep1} and \emph{ra\_rep2}, which illustrate a common problem when faced with real data set: the lack of observations. 
Indeed, initial assumptions for using LAD or MCP are that experts should be able to provide enough data to learn something relevant.

The fact that the instances do not follow a uniform probability distribution does not invalidate our probabilistic model.
When the reduction is applied by computing a minimal $\Y$, the behaviour of the data set evolves due to the fact that we reach complete Boolean functions. 
This will be considered in Section \ref{sec:scope}.


\section{Discussion and Evaluation of our Results}
\label{sec:scope}
    

Logical Analysis of Data enables a better understanding of the structure of groups of binary data; in particular logical patterns and formulae can be used to characterize a group of positive observations against a group of negative observations. 
Nevertheless, when faced to practical instances, complete search algorithms are subject to computational difficulties, for example when practitioners aim at examining complete sets of patterns of logical formulae for making decision.
Moreover, once characterizations have been proposed to users, several questions may arise in terms of robustness and covering. 
These difficulties have been pointed out in seminal works on LAD and further explored, e.g., in \cite{Hammer2006,Chhel2013,ChambonBLS18}.
We introduced several probabilistic models in this paper to further investigate such questions; through them, we were able to address several important problems as surveyed in Section~\ref{sec:overview}.

\subsection{Contributions}
We believe that the main contribution of the present work is a mathematical modeling that can provide a new quantitative formalization for logical data analysis. 
We have also shown how such an approach can be used to analyze real instances with uniformly distributed observations, and is an important computational tool for assessing the learning capabilities of the proposed methods. 
Of course, we do not exhibit the performances of learning algorithms here; but since these algorithms are based on Branch and Bound exploration methods, it is clear that reducing the initial bounds will lead to an improvement of these performances. 

Our model can also be used to compute possible evolutions of the instances if new observations are introduced. 
We have experimented with this aspect but, for sake of concision, we do not report the data here; however the results are coherent with the different behaviours of instances already reported. 

Moreover, our probabilistic approach allows for a better understanding of why some instances could not be reduced or explained by simple patterns. 
In such cases, it seems that the initial data groups provided by practitioners do not have the expected properties that would allow us to learn some relevant characterizations. 

A probabilistic approach based on the assumption of uniform distribution of data also allows us to clearly identify data groups that are not close to uniform distributions (when the probability of obtaining real computed results is low) and thus groups that contain relevant information. Such an approach can be used as a test for classifying data groups. 

Finally, once solutions have been computed, the covering power is an important question. 
Again, a probabilistic approach might be relevant to evaluate whether using more attributes could reasonably lead to more robust solutions (even if suboptimal). 
The nature of the data sets change since the underlying Boolean function become more precise after the reduction process, when applicable.

\subsection{Effective computations}
The actual computations of the probabilities presented in this paper were done using the Computer Algebra System  ({\small CAS}) Maple, and were quick enough that there was no need to turn to asymptotics.
The most complicated of our computations appear in Model M2: they are relative to the probability that the intersection of projections has a size bounded by some value~$r$; for $r=4$ it was roughly $0.1$ second.

We point out that, once the mathematical expressions for the desired probabilities have been established, the numerical computations do not rely on specific features of some {\small CAS}, but are a simple implementation of recurrence relations given in Section~\ref{sec:appendix}: as such, they can be easily implemented in any computing language and integrated into an existing solver.

\subsection{Probabilistic assumptions}
We made two important assumptions on our data: the attributes are independent, and the observations are independant and drawn according to a uniform distribution from the group of all possible observations.
Such assumptions are standard in the analysis of algorithms: the first modelization is done under uniformity and independance assumptions; once the model and its associated computations have been established and solved, feedback from actual data suggests in which direction the model should be extended.
The present paper is at the first stage; its aim is to show what analysis of algorithms can bring to LAD, and more generally to Constraint Programming.
Our tools allow us to consider non-uniform distributions or dependent attributes, by adjusting the generating functions (the dependency between observations might require a further model).
Of course, the first step would be to characterize the non-uniform distributions or dependency assumptions that we wish to consider. 
If we can sum up this information as probability distributions on the sets of aggregated attributes $\Y$ and~$\Z$, then it is possible to encode this in the generating functions.
Expressing the desired probabilities through coefficients of these functions remains unchanged; then extracting the coefficients, at least asymptotically, might fall within the full power of Analytic Combinatorics tools.

\medskip
When dealing with real data, we often cannot know a priori whether our probabilistic assumptions are valid. But, in cases where we already have solutions (e.g., given by \Patterns\ or \MCP), we can compare these results with our theoretical results; this allows us to learn whether these assumptions seem to hold, or are widely off.

The most important point is to recall that LAD aims at extracting a more compact representation of a hidden partially defined Boolean function induced by the partitioned data set into $P$ and $N$. 
When reducing the number of attributes by using \MCP~or \Patterns, the nature of the function changes and it may become fully defined (i.e., all possible observations, after projection on $\Y$, are present in the data set). 
\MCP\ tends to reduce as much as possible $|\Y|$.
When using patterns the idea is to get a full cover of $P$, i.e., to find a Boolean function that is fully defined on $P$ even if some of its values are missing on $N$.

From these considerations, some conclusions may be draw to assess the possible benefits of the probabilistic model we propose here. These conclusions are of course supported by computations on the data sets we had.
\begin{itemize}

    \item If there is a huge number of observations issued from a uniform distribution, it is impossible to reduce the number of attributes to extract some kind of knowledge. 
    This has been confirmed by experiments (not reported here) on random instances.
     
    \item
    If we have too few observations compared to the number of attributes that describe them, it is difficult to obtain a small set $\Y$; hence the data set may behave similarly to uniformly distributed observations. 
    Nevertheless our computations show that the resulting reduction of the number of attributes, even when small, allows us to argue for non-uniformly distributed observations.  
    
    \item
    When the reduction process leads to a fully defined Boolean function, then due to the fact that all possible observations are present, our probabilistic model shows interesting estimations. 
    As previously mentioned, this is of course the most interesting case for practitioners: it provide them with a full diagnosis process. 
    
    \item 
    If the reduction process leads to partially Boolean functions, the resulting data set does not exactly fit with our assumptions. 
    Nevertheless, the probabilistic computations may be still valid to estimate rough bounds for $|\Y|$. 
    
\end{itemize}

\subsection{Asymptotics}

A reader familiar with Analytic Combinatorics might remember that its full power comes when considering a fixed structure for some objects or data, and letting their size go to infinity.
In our present approach, the structure of our data, hence the generating function from which we compute the coefficients relevant to our problem, is not fixed: it depends on the sizes of the attribute sets~$\Y$ and~$\Z$.
This leads to a multi-valued problem and it is not clear at first glance which parameters: sizes of domains, of observation sets, or of projections, go to infinity, and what are their respective orders of growth.
But the very meaning of our data and of the information we are trying to extract from it provides a way: we wish to minimize $|\Y|$, i.e., to maximize~$|\Z|$; when the size of the instance and the number of attributes are ``large enough'', this actually allows us to assume that $|\Z|$ and, even more so, $d_\Z = 2^{|\Z|}$ are infinite.
Then a simple change in the generating functions that we use to describe the instances takes us back to the standard framework of Analytic Combinatorics.
Section~\ref{sec:calculs-asympt} presents a mathematical justification of this change and shows on examples how we can still extract information for the modified generating functions.
It also makes clear that, to deal with this multi-valued problem, the most important part might be deciding on the respective orders of growth of the different parameters (sizes) in the initial situation.

\subsection{Possible extensions}
In the framework of the problems considered in the present paper, it may be desirable to consider three or more groups.
Our model allows for this, but the generating functions involved in describing the possible situations become more complex.
This will probably lead to longer computation times, which would restrict the applicability of our approach: in such a situation we would probably have to turn to asymptotics for efficient computations.

A possible extension might be to consider non-uniform distributions on data, with a first choice being a distribution of the so-called 80\%-20\% type.
The generating functions are easy to adapt to such a situation and lead to manageable computations; see for example~\cite{gardy-nemirovsky} on a similar model for database projections.

On the practical size, we may also experiment with the possible improvements that could be obtained for the standard algorithms for solving LAD related problems. 
For instance, the bounds we compute could be easily introduced in Mixed Integer Linear Programming models.

A long-range goal of this paper was to show how the methods of Analytic Combinatorics can help in analyzing the performance of various algorithms in LAD and related fields. 
In this optic, our paper is in the same vein as the earlier approach presented in~\cite{BGLT2013}, which was concerned with the analysis of an algorithm in Constraint Programming ({\sc AllDifferent}).
We believe that our present work proves again the interest of establishing a link between these two fields of research.

\medskip
{\bf Acknowledgments.}
DG thanks Brigitte Vall\'ee for stimulating discussions on a former version of this paper.


\section{Appendix: computation of generating functions}
\label{sec:appendix}


\subsection{Analytic Combinatorics and Enumeration}
\label{sec:intro-AC}

Consider a class $\C$  of objects to which we associate a size, i.e. a natural number that we assume to be finite, and define $c_n$ as the number of elements of~$\C$ with size~$n$.
We can define a \emph{generating function} associated with the class~$\C$ by $C(z) = \sum_{n \geq 0} c_n z^n$.
This generating function encapsulates all that we know about the numbers~$c_n$, and knowing the sequence $\{ c_n, n \geq 0 \}$ is (theoretically) equivalent to knowing their generating function~$C(z)$.

Now getting the exact values for the $c_n$ can be cumbersome, and the function $C(z)$ is often easier to obtain.
Of course, this means that we must recover the numbers~$c_n$ from the function~$C(z)$; we write $c_n = [z^n] C(z)$, meaning that $c_n$ is the coefficient of $z^n$ in~$C(z)$, now considered as an entire series.

A naive reader might wonder: Why go through the trouble of writing down generating functions, then extracting their coefficients, instead of just considering the numbers~$c_n$?
The answer is twofold: in many cases Analytic Combinatorics offers technics both for getting the generating functions in a systematic way, and for asymptotic evaluation of the coefficient, when extracting its exact value is too complicated or not informative of its behaviour.

\medskip
We illustrate this on the classical case of binary planar trees, sometimes called Catalan trees.
This class $\cal C$ contains the the empty tree and the trees built by taking a node and adding two trees as the left and right subtrees.
It satisfies the recurrence relation
\[
{\cal C} = \{ \oslash \} \cup ( \bullet, {\cal C}, {\cal C} ).
\]
Defining the size of a tree as the number of its nodes (the empty tree is the only tree of size~0) and setting $C(z) = \sum_n c_n z^n$ with $c_n$ the number of trees of size~$n$, the symbolic method readily gives
\[
C(z) = 1 + z \, C(z)^2 = \frac{1-\sqrt{1-4z}}{2z} 
\]
from which we obtain $c_n ={ \binom{2n} {n}} / (n+1)$.
The asymptotic value $c_n \sim 4^n \ n\sqrt{\pi \, n}$ can be either obtained from the exact expression by Stirling's factorial approximation, or directly extracted from the expression of $C(z)$ by singularity analysis.

\medskip
The scope of Analytic Combinatorics is of course much larger that can be considered in this quick introduction and we refer the reader to the literature on the subject for more information on Analytic Combinatorics, most notably to the two books by Flajolet and Sedgewick \cite{fs96,livre-flajolet}.

We now turn to the counting of instances in the next Section, before taking into account groups in Section~\ref{sec:complete-gf} and the models M1 and M2 in further Sections.

\subsection{Generating Function for Instances with two Attributes}
\label{sec:instances-2attributs}

This section is a reformulation and adaptation of some of our former studies on the sizes of projections in relational databases; see for example~\cite{gardy-coeffs,gardy-survey-urnes}.

\paragraph{Instances as sets of values}
Let us first consider instances as sets of observations, without further knowledge on their structure.
The observations are drawn from a domain of possible values of size~$d$, and any value either belongs to the instance, or not.
Hence the (very simple) generating function describing whether a specific value belongs to the instance is $1+z$, with $1$~corresponding to the absence and $z$ to the presence of that value in the instance.

Consider next the $d$ possible values, whose presences in the instance are independent: this is described by multiplying the simple generating functions associated to each value, and we obtain $(1+z)^d$. 
Taking the coefficient of $z^n$ in this function, we obtain ${d \choose n}$, which is of course the number of possible instances with $n$ values.

The reader should note that our independence assumption on values does not contradict the fact that we shall be investigating instances with~ $n$~values: the global generating function counts instances of all possible sizes, then we recover the constraint on the size~$n$ by extracting the coefficient of the $n$-th power of~$z$.

\paragraph{Introducing the attributes}

We now refine this basic model to take into account the existence of two (sets of) attributes for the instances; we call them~$\Y$ and~$\Z$.
We want to count the instances according to their size and their number of distinct values on the attribute~$\Y$: there are now two parameters (two sizes) and we need two variables to ``mark'' them.
We use the notation $\pi_\Y$ (projection on the attribute~$\Y$) for the number of distinct values of~$\Y$ that appear in the instance.

When considering instances without attributes, we first wrote the generating function describing what happens for a single value, then raised this function to the $d$-th power, $d$ being the number of possible values.
Here we follow a similar approach, but now the values we consider are those of the domain of the attribute~$\Y$.
Each of those values either does not appear in the instance, which we mark by~1, or does appear in one or more observations. 
Let us use the variable~$z$ to mark these observations by~$z$; then the set of all possible non-empty values of~$\Z$ associated to a fixed value of~$\Y$ is described by the function $(1+z)^{d_Z} -1$.
Now use the variable~$y$ to mark the specific value of~$\Y$ we have been considering: the set of observations that contain this value is described by the function $ 1 + y \left( (1+z)^{d_Z} -1 \right) $. 
Finally, through the independence assumption for observations, we raise the last expression to the $d_\Y$-th power to obtain the generating function $f(y;z)$ enumerating all possible values of an instance, with $z$ marking its size and $y$ the size of $\pi_\Y $ as
\begin{equation}
\label{eq:premiere_def_f}
f(y;z) = \left( 1 + y \left( (1+z)^{d_Z} -1 \right)  \right)^{d_\Y} = \sum_{k,n \geq 0} f_{k;n} \; y^k z^n,
\end{equation}
where $f_{n;k}$ is the number of instances of size $n$ such that their projection on $\Y$ has size~$k$.
Preparing the work for what follows, we rewrite the function $f(y;z)$~as $(1 + y \phi (z) )^{d_y}$ with
\begin{equation}
\label{eq:def-phi}
\phi(z):=  (1+z)^{d_\Z} -1 = \sum_{n \geq 1} \binom{d_\Z}{n} z^n .
\end{equation}
It should be noted that the function $\phi(z)$ actually depends on the domain size~$d_\Z$ and ultimately on the number $|\Y|$ of attributes in the set~$\Y$, although we do not make explicit this dependency in our notations, to keep them as simple as possible.

From Equations~\eqref{eq:premiere_def_f} and \eqref{eq:def-phi}, we obtain $f_{k;n} = \binom{d_\Y}{k} \; \alpha_{k;n} $
with
\begin{equation}
\alpha_{k;n}:= [ z^n ]  \phi(z)^k    = \sum_{r} (-1)^{k-r} \; \binom{k}{r} \;  \binom{r d_\Z}{n} .
\label{eq:def-alpha}
\end{equation}
We can begin the summation at~1 as the binomial coefficient $\binom{r d_\Z}{n}$ is null when $r=0$;  we also have that $\alpha_{0,0} = 1$  and $\alpha_{k;n} = 0$ for $k=0$ and $n \geq 1$ or for $n<k$, i.e., $\alpha_{k;n} >0$ for $k=n=0$ or for $1 \leq k \leq n$.

The total number of instances with $n$ distinct couples is $\sum_k f_{n;k}  = \binom{d_\Y \, d_\Z}{n}$, which shows that $\binom{d_\Y \, d_\Z}{n} =  \sum_k \binom{d_\Y }{k} \, \alpha_{k;n}$.

\medskip
For the record, we note the parallelism between our numbers $\alpha_{n;k}$ counting the number of ways to partition $n$ couples $(y,z)$ of $D_\Y \times D_\Z$ into $k$ blocks of couples sharing a common value on the attribute~$\Y$ (with $k$ the number of different values on~$\Y$), and Stirling numbers of the second kind counting the number of ways to partition $n$ elements into $k$ non-empty sets: 
\[
\stirling{n}{k} = \left[ \frac{x^n}{n!} \, y^k \right] e^{y (e^x -1)}
= \sum_{r \geq 0} (-1)^{k-r} \binom{k}{r} \frac{r^n}{k!}.
\]

\paragraph{Probability that an instance has a projection of given size}

The conditional probability $Pr(k/n)$ that an instance has a projection of size~$k$, knowing the size $n$ of the instance (and assuming that all instances are equally likely), is simply the number $f_{k;n}$ of such instances divided by the total number $\binom{d_\Y \, d_\Z}{n}$ of instances of size~$n$:
\begin{equation}
Pr(k/n) =  \frac{[y^k z^n ] f(y;z)}{[y^n] f(1;z)}
= \frac{\binom{d_\Y}{k}} {\binom{d_\Y \, d_\Z}{n}} \, \alpha_{k;n} .
\label{eq:proba-one-group}
\end{equation}
This allows for a quick computation of this probability;  if relevant, we can also consider its asymptotic behaviour for large domain and instance sizes. 
E.g., former studies have shown that the normalized distribution of the projection size is often asymptotically Gaussian~\cite{gardy-puech84}.

	\subsection{Groups and their Generating Function}
	\label{sec:complete-gf}

In this section we further complexify our model and work out the generating function that captures all the information about the two groups $G_1$ and $G_2$ of an instance with two attributes~$\Y$ and~$\Z$. 
This requires us to introduce new variables, in order to take into account the additional information.
By giving specific values to some of those variables, we can then obtain from the new, global generating function the  specific generating functions for various models M1 and M2 and cases {\bf (A)} to {\bf (F)} (see Table~\ref{table:differents-cas}).
The variables we shall use are the following ones:
\begin{itemize}
	\item[--] $z$ marks the total number of observations in the instance and $z_i$ ($i = 1,2$) the observations in the group $G_i$;
	\item[--] $t$ marks the observations that belong to both groups, i.e., the intersection $G_1 \cap G_2$;
	\item[--] $y$ marks the  projection of the whole instance on $\Y$, i.e., the different values taken by the observations on the attribute~$\Y$;
	\item[--] the $y_i$ ($i = 1,2$) mark the projections $\pi_\Y (G_i)$, $x$ for $\pi_\Y (G_1) \cap \pi_\Y (G_2)$;
	\item[--] finally, $x$ marks the intersection of the projections $\pi_\Y (G_1) \cap \pi_\Y (G_2)$.
\end{itemize}

Closely paralleling the reasoning in Section~\ref{sec:instances-2attributs}, we first compute the function $\tilde f (x; y_1, y_2; t ; z_1,z_2)$ describing the possibilities for a fixed value~$a$ of~$D_\Y$ to appear in an instance.
We then raise this function to the $d_\Y$-th power to obtain the global generating function $\tilde F (x; y_1, y_2; t ; z_1,z_2) $ counting all the possible cases of group and projection sizes.

Let us first build the function $\tilde{f} (x; y_1, y_2; t ; z_1,z_2)$ by considering the possible situations for a value~$a$ of~$D_\Y$
\begin{itemize}
    \item[--] If $a$ does not appear in the instance $G$, we mark this with the term~1.
    
    \item[--] If $a$ appears in $G$, then $a \in \pi_\Y (G)$. 
    Assume first that $a$ appears in $\pi_\Y (G_1)$ but not in $\pi_\Y (G_2)$.     Marking the sizes of $G_1$ and $\pi_\Y(G_1)$ respectively by $z_1$ and $y_1$ gives the term $y_1 \, \phi(z_1)$. The symmetric case $a \in \pi_\Y (G_2)$ but $a \not \in \pi_\Y (G_1)$ gives the term $y_2 \, \phi(z_2)$.
    
    \item[--] If $a$ appears in $\pi_Y (G_1) \cap \pi_\Y (G_2)$, this can happen because (at least) one couple $(a,b)$ belongs to $G_1 \cap G_2$, or because (at least) one couple $(a,b)$ belongs to $G_1$ but not to $G_2$ and another couple $(a,c)$ belongs to $G_2$ but not to $G_1$, or because both situations happen: (at least) one couple $(a,b)$ belongs to $G_1 \cap G_2$, another couple $(a,c)$ belongs to $G_1$ but not to $G_2$, and yet another couple $(a,d)$ belongs to $G_2$ but not to~$G_1$.
    
    Now define a partition of $D_\Z$ according to~$a$, as follows: $E_0(a)$ is the set of values $b \in D_\Z$ such that $(a,b) \not \in G$, $E_1(a)$ the set of values s.t. $(a,b)  \in G_1$ but $(a,b) \not \in G_2$ and similarly $E_2(a)$, and finally $E_{12}$ is the set of values $b$ s.t. $(a,b) \in G_1 \cap G_2$.
    The expressions associated to these subsets are respectively $1$ for $E_0$, $z_i$ for $E_i(a)$, and $t \, z_1 \, z_2$ for $E_{1,2}$ and we can write down the generating function marking the different possibilities for a specific value $b \in D_Z$,  where the variable $t$  marks  the (possibly empty) intersection of $G_1$ and $G_2$; this is
    \[
    1 + z_1 + z_2 + t z_1 z_2 .
    \]
    By raising this function to the power~$d_\Y$ to take into account all the values of $D_\Y$, then substracting~1 to avoid the case where $\pi_\Y (G)$ is empty (which cannot happen in our context), we get the generating function encapsuling the information on all the couples $(a, \star)$ as:
    \[
    (  1 + z_1 + z_2 + t z_1 z_2 )^{d_\Z} -1 =
    \phi (z_1 + z_2 + t z_1 z_2 ).
    \]
    We introduce now the variable $x$ to mark  the (non-empty) intersection $\pi_\Y (G_1) \cap \pi_\Y (G_2)$, the information on the projection is described by the factor $x y_1 y_2$, which we multiply by the expression describing the groups in the situation we are currently exploring, namely
    \[
    \phi(z_1 + z_2 + t z_1 z_2) - \phi(z_1) - \phi (z_2),
    \]
    where the substraction of the terms $\phi(z_1)$ and $\phi(z_2)$ comes from  excluding the cases where $a$ belongs to exactly one of the groups $G_1$ or~$G_2$, which is already considered.
\end{itemize}
Putting all this together gives the generating function describing the allowed set of tuples $(a,\star)$ for any given value $a \in D_\Y$:
\begin{eqnarray*}
&& \tilde f (x; y_1, y_2; t ; z_1,z_2) =
\\
&& 1 + y_1 \, \phi(z_1) + y_2 \, \phi(z_2) + x \, y_1 \, y_2 \, \left[ \phi(z_1 + z_2 + t z_1 z_2) - \phi(z_1) - \phi(z_2) \right].
\end{eqnarray*}
By considering all the possible values of $D_\Y$, we finally obtain the full generating function as
\begin{equation}
\label{eq:fg-totale}
\tilde F (x; y_1, y_2; t ; z_1,z_2) = 
\tilde f (x; y_1, y_2; t ; z_1,z_2)^{d_\Y}.
\end{equation}

\medskip
We are now in a position to write the generating functions for the different models of Section~\ref{sec:empty-or-not}. 
\begin{enumerate}
    \item 
Setting $x=0$ in the expression of $\tilde{F}$ gives the generating function describing the instances with two groups $G_1$ and $G_2$ such that $\pi_\Y (G_1) \cap \pi_\Y (G_2) = \emptyset$.
This is model M1($\Y $); see Section~\ref{sec:two-groups-models-I}.

\item
If we keep the variable $x$ but set $t=0$ in $\tilde F$, we obtain the generating function for the model M2($\Y$), where $G_1 \cap G_2 = \emptyset$ but the intersection $\pi_Y (G_1) \cap \pi_\Y (G_2)$ may be non-empty; see Section~\ref{sec:two-groups-models-II}.

\item
Finally working with the full generating function $\tilde F (x; y_1, y_2; t ; z_1,z_2) $ might correspond to Model M3 in which the two groups are not disjoint, although we considered here a specific partition of the attribute set $\X$ into~$\{ \Y, \Z \}$: it would be more relevant to write a direct description of this case that does not involve a partition of the set of attributes.

\end{enumerate}

	\subsection{Generating Functions for Model M1($\Y$)}
	\label{sec:two-groups-models-I}

Here we consider instances that are partitioned into two groups $G_1$ and~$G_2$, in such a way that the intersections $G_1 \cap G_2$ and $\pi_\Y (G_1) \cap \pi_\Y (G_2)$  are both empty.
We have seen in the preceding section that the generating function describing such instances is 
\begin{equation}
\label{eq:fg-globale-modeleI}
F(y_1,y_2; z_1,z_2) =  \Tilde{F}(0; y_1,y_2; 0; z_1,z_2 ) =( 1+ y_1 \phi(z_1) + y_2 \phi(z_2) )^{d_\Y}.
\end{equation} 
The variable $t$ does not matter here, as it can only be present in terms where $x$ also appear, so we might as well set it to~0.

\subsubsection{The different cases}
\label{sec:appendice_differentscas_M1}
We next consider how Equation~\eqref{eq:fg-globale-modeleI} can be specified according to the information we have on the sizes of the instance, its groups, and their various projections on attribute~$\Y$ (see table~\ref{table:differents-cas}).

\paragraph{Case {\bf (A)}: When we only know the size $n$ of the instance}
In practical cases, we always assume that $n \geq 1$.
This corresponds to setting $y_1=y_2=1$ and $z_1=z_2=z$ in the equation~\eqref{eq:fg-globale-modeleI}; we obtain
\[
F (1,1;z,z) = (1+2\phi(z))^{d_\Y} =  \left( 2 (1+z)^{d_\Z} - 1 \right)^{d_\Y}
\]
and the number $\rho_{n}:=  [ z^n ] F (1,1;z,z) $ of such instances is
\begin{equation}
\label{eq:nombre-total}
\rho_{n}  = \sum_{1 \leq k \leq n} 2^k \binom{d_Y}{k} \, \alpha_{k;n}.
\end{equation}
This is the normalization factor that we shall use for computing probabilities of various situations, when we only know the total size of the instances and that its observations are divided into two non-empty groups $G_1$ and~$G_2$, without further information on sizes.

\paragraph{Case {\bf (B)}: When we know the sizes $n_1$ and $n_2$ of the groups $G_1$ and $G_2$}
Here again we assume that $n_1, n_2 \geq 1$.
\[
F (1,1;z_1,z_2) = (1+\phi(z_1) + \phi(z_2) )^{d_\Y} =  \left( (1+z_1)^{d_\Z} + (1+z_2)^{d_\Z} - 1 \right)^{d_\Y}.
\]
The number of instances is now $\rho_{n_1, n_2} =[z_1^{n_1} \, z_2^{n_2}] F(1,1; z_1, z_2)$ which readily gives
\begin{equation}
\label{eq:nbre-total-n1-n2}
\rho_{n_1,n_2}  = \sum_{1 \leq k_1 \leq n_1, 1 \leq k_2 \leq n_2} \binom{d_Y}{k_1, k_2} \, \alpha_{k_1;n_1} \; \alpha_{k_2;n_2}.
\end{equation}
Of course $\rho_n = \sum_{n_1, n_2 \geq 1 ; n_1 + n_2 = n} \, \rho_{n_1,n_2}$.
We shall use this normalization factor when computing probabilities in situations where the sizes of the two groups are known.

\paragraph{Case {\bf (C)}: When we know the sizes $n$ of $R$ and $k$ of its projection $\pi_\Y (R)$}

Here $1 \leq k \leq n$.
The function describing this situation is simply 
\[
F (y,y;z,z)  = [ y^k z^n ] (1+2 y \phi(z))^{d_\Y}
\]
 and the number of instances satisfying the size conditions  becomes $\beta_{k;n} = [ y^k z^n ] F (y,y;z,z)$; hence
\[
\beta_{k;n} = 2^k \binom{d_\Y}{k} \, \alpha_{k;n}.
\]
We check that $ \rho_n = \sum_k \beta_{k;n}$.
The probability that a random instance of size~$n$ has a projection of size~$k$~is
\[
Pr(k/n) = \frac{\beta_{k;n}}{\rho_n} .
\]

\paragraph{Case {\bf (D)}: When we know $n$ and the sizes $k_1$ and $k_2$ of the group projections}

We assume here that $k_1, k_2 \geq 1$ and $n \geq 2$ (there is at least one observation in each group $G_i$).
This case is described by the generating function $F(y_1,y_2;z,z) = ( 1+ (y_1+y_2) \phi(z) )^{d_\Y}$; we take the coefficient of $y_1^{k_1} y_2^{k_2} z^n $ in it.
The number of such instances involves multinomial coefficients:
\[
\lambda_{k_1,k_2;n}:= \binom{d_\Y}{k_1 , k_2} \, \alpha_{k_1+k_2,n}.
\]
Taking into account the equality $ \sum_{ k_1 + k_2 = k} \, \binom{d_\Y}{k_1 , k_2} \, \alpha_{k;n} = 2^k \, \binom{d_\Y}{k}$, we expect that $\beta_{k;n} = \sum_{k_1+k_2 = k} , \lambda_{k_1,k_2;n}$.
But in this sum $k_1$ or $k_2$ may be null, and the valid equality is $\beta_{k;n} = \sum_{k_1, k_2 \geq 1; k_1+k_2 = k} \, \lambda_{k_1,k_2;n}$.

The probability that the groups have projections of sizes $k_1$ and $k_2$, knowing the total size $n$ of~$R$,~is
\[
Pr(k_1, k_2 / n) = \frac{\lambda_{k_1,k_2;n}}{\rho_n}.
\]

\paragraph{Case {\bf (E)}: When we know the group sizes $n_1$, $n_2$ and the total size $k$ of the projection}

Here $n_1, n_2 \geq 1$ and $2 \leq k \leq n_1 + n_2$ (the projections of the groups $G_1$ and $G_2$ on $\Y$ are both non-empty, hence $k=1$ is impossible).
We work with 
\[
F(y,y; z_1,z_2) = (1 + y (\phi(z_1) + \phi(z_2)))^{d_\Y}
\]
  and take the coefficient of $y^{k} z_1^{n_1} z_2^{n_2} $ in it.
  Again assume that $n_1,n_2 \geq 1$, which implies that $k_1,k_2 \geq 1$.
The number $\gamma_{k;n_1,n_2}$ of instances with group sizes $n_1$ and $n_2$, and total projection size~$k$, is
\[
\gamma_{k;n_1,n_2} = \sum_{k_1 + k_2 = k} \binom{d_\Y}{k_1,k_2} \, \alpha_{k_1,n_1} \alpha_{k_2,n_2}.
\]
We check that $\rho_{n_1,n_2} = \sum_{k=1}^{d_\Y} \gamma_{k;n_1,n_2}$.
The probability that the projection size is $k$, knowing the sizes $n_1$ and $n_2$,~is
\[
Pr(k/n_1,n_2) = \frac{\gamma_{k;n_1,n_2}}{\rho_{n_1,n_2}}  .
\]

\paragraph{Case {\bf (F)}: When we have full information on sizes}

Now we know $n_1$, $n_2$, $k_1$ and $k_2$, which are such that $1 \leq k_1 \leq n_1$ and $1 \leq k_2 \leq n_2$.
We need the coefficient of $y_1^{k_1} y_2^{k_2} z_1^{n_1} z_2^{n_2}$ in the whole function $F(y_1,y_2;z_1,z_2)$.
This is
\[
\delta_{k_1,k_2;n_1,n_2} = \binom{d_\Y}{k_1,k_2} \, \alpha_{k_1;n_1} \alpha_{k_2;n_2}.
\]
Of course
$\sum_{k_1 + k_2 = k} \delta_{k_1,k_2;n_1,n_2} = \gamma_{k;n_1,n_2}$
and 
$\sum_{n_1 + n_2 = n}\delta_{k_1,k_2;n_1,n_2} = \lambda_{k_1,k_2;n}$.
The probability that the projections of the groups $G_1$ and $G_2$ have sizes $k_1$ and~$k_2$, knowing that the groups themselves have sizes $n_1$ and~$n_2$,~is
\[
Pr(k_1,k_2 / n_1, n_2) = \frac{\delta_{k_1,k_2;n_1,n_2} }{\rho_{n_1,n_2}} .
\]

\subsubsection{Summary of results for Model M1($\Y$) and effective computation}

For quick reference, Table~\ref{table:questions-vs-coeffs_bis}\footnote{This is Table~\ref{table:questions-vs-coeffs} extended to include the relevant generating function coefficients.} sums up the questions that are relevant from an algorithmic point of view, the probabilities we want to compute, and the generating function coefficients involved; then
Table~\ref{fig:M1-coefficients} gives expressions for these coefficients.

\begin{table}[h]
\begin{center}
 
 \begin{tabular}{|c|c|c|}
\hline
Question & Probability  & Coefficients \\
\hline \hline
  Reasonable bound for $|\Y|$ & $Pr(n_1,n_2 / n  ) $ & ${\rho_{n_1,n_2}} /{\rho_n}$    \\ 
 \hline
  Analysis of  $|\Y|$  & $Pr (k_1,k_2 / n_1, n_2)$  & ${\delta_{k_1,k_2; n_1,n_2}} /{\rho_{n_1,n_2}}$ \\ 
   \hline
  Analysis of the covering of  $\Y$  & $Pr (k / n_1, n_2)$  & ${\gamma_{k; n_1,n_2}}/{\rho_{n_1,n_2}}$ \\ 
  \hline 
  Reliability of size  $|\Y|$  & $Pr (k=n / n_1, n_2)$  & ${\gamma_{n; n_1,n_2}}/{\rho_{n_1,n_2}}$ \\ 
  \hline
  Existence of unique pattern & $Pr(k_1=1 / n_1, n_2)$  & ${\sum_{k_2} \delta_{1,k_2; n_1,n_2}}/{\rho_{n_1,n_2}}$ \\ 
  \hline
 \end{tabular}
 
\end{center}

\caption{\label{table:questions-vs-coeffs_bis} 
Probabilities or Model M1($\Y$),  including the generating function coefficients that answer these questions. }
\end{table}

\begin{figure}[h]
\begin{center}
\begin{tabular}{|l c|l c|}
\hline
  {\bf (A})& $\rho_n = \sum_k 2^k \binom{d_\Y }{k} \alpha_{k;n} $ & {\bf (B})& $\rho_{n_1,n_2} = \sum_{k_1,k_2} \binom{d_\Y}{k_1,k_2} \alpha_{k_1;n_1} \alpha_{k_2;n_2} $ \\
\hline 
&&& \\
 {\bf (C})&  $\beta_{k;n} =  2^k \, \binom{d_\Y}{k} \, \alpha_{k;n}$ & {\bf (E})&  $\gamma_{k;n_1,n_2} =\sum_{k_1 + k_2 = k} \binom{d_\Y}{k_1,k_2} \alpha_{k_1;n_1} \alpha_{k_2;n_2} $ \\
\hline  
&&&\\
&&&\\
 {\bf (D})& $\lambda_{k_1,k_2;n} =  \binom{d_\Y}{k_1, k_2} \, \alpha_{k_1 + k_2; n} $  & {\bf (F})&  $\delta_{k_1,k_2;n_1,n_2} = \binom{d_\Y}{k_1, k_2} \, \alpha_{k_1;n_1} \alpha_{k_2;n_2}$ \\ 
\hline 
\end{tabular}

\end{center}
\caption{\label{fig:M1-coefficients}
Expression for coefficients in Model  M1($\Y$) with $\alpha_{k;n} = \sum_{r} (-1)^{k-r} \; \binom{k}{r} \;  \binom{r d_\Z}{n}$. }
\end{figure}

For effective computation, we should first compute the numbers $\alpha_{k;n}$, then  $\beta_{k;n}$, $\lambda_{k_1,k_2;n}$ and $\delta_{k_1,k_2;n_1,n_2}$ as they are easily obtained by multiplying a term $\alpha_{k;n}$ or the product of two such terms by a binomial or multinomial coefficient.
Next we can compute $\rho_n$ as the sum of the $\beta_{k;n}$ over~$k$ , and $\gamma_{k;n_1,n_2}$ as the sum  of the $\delta_{k_1,k_2;n_1,n_2}$ over all $k_1$ and $k_2$ such that $k_1+k_2 = k$.
Finally $\rho_{n_1,n_2}$ is obtained as the sum of the~$\gamma_{k;n_1,n_2}$ over~$k$, which is none other than the formula coming from Equation~\ref{eq:nbre-total-n1-n2} when we decompose it into sub-summations.

	\subsection{Generating Functions for Model  M2($\Y$)}
	\label{sec:two-groups-models-II}

In this section we enumerate the instances that are partitioned in two groups $G_1$ and $G_2$  but without any condition on the projections $\pi_\Y (G_1)$ and $\pi_\Y(G_2)$: the condition $G_1 \cap G_2 = \emptyset$ holds but the intersection $\pi_\Y (G_1) \cap \pi_\Y(G_2)$ may be $\neq \emptyset$.
The relevant generating function is obtained by setting $t=0$ in Equation~\eqref{eq:fg-totale} of Section~\ref{sec:complete-gf} and we get
\begin{equation}
\label{eq:fg-casM2}
 F (x; y_1, y_2; 0 ;z_1,z_2) = 
 f (x; y_1, y_2; 0; z_1,z_2)^{d_\Y}
 \end{equation}
 where $ f (x; y_1, y_2; 0; z_1,z_2) $
 is now simply
 \[
1 + y_1 \, \phi(z_1) + y_2 \, \phi(z_2) + x \, y_1 \, y_2 \, \left[ \phi(z_1 + z_2 ) - \phi(z_1) - \phi(z_2) \right].
\]

Just as we did for Model M1($\Y$), it is possible to work out the generating functions for cases {\bf (A)} to {\bf (F)}.
However the numerical examples we present in Section~\ref{sec:ModelM2-calculs} only require the knowledge of the numbers $\rho_{n_1,n_2}$, which we compute below, and of the size of the intersection $\pi_\Y (G_1) \cap \pi_\Y(G_2)$, which we address in Section~\ref{sec:M2-size-intersection}.

\subsubsection{Number of instances with given sizes of groups}

This is our Case {\bf (B)}: we know the sizes $n_1$ and $n_2$ of the groups $G_1$ and $G_2$.

The generating function counting all possible configurations of groups $G_1$ and $G_2$ is now obtained by setting $y_1 = y_2 = x =1$; as expected, this is simply
\[
F(1; 1,1; 0;z_1,z_2) = ( 1 + z_1 + z_2 )^{d_\Y d_\Z} ,
\]
and the number of instances that have exactly $n_1$ elements in $G_1$, $n_2$ elements in~$G_2$, and none in~$G_1 \cap G_2$ is the coefficient of $z_1^{n_1} \, z_2^{n_2}$:
\[
\rho_{n_1,n_2} = \binom{d_\Y d_\Z}{n_1, \, n_2}.
\]
Here again this formula can be derived directly; requiring that $n_1, n_2 \geq 1$ will ensure that both groups are non-empty.

\subsubsection{Size of the Intersection of Projections}
\label{sec:M2-size-intersection}

We consider here the probability that the intersection $\pi_\Y (G_1) \cap \pi_\Y (G_2)$ has size~$u$. 
As we only consider the case {\bf (B)} where we know the sizes $n_1$ and $n_2$ of the groups $G_1$ and $G_2$, the appropriate generating function~is
\[
F(x; 1,1;0;z_1,z_2) =
\left( 1 + \phi (z_1) + \phi(z_2) + x [ \phi(z_1+z_2) - \phi(z_1) - \phi(z_2)  ]  \right)^{d_\Y}.
\]
Taking the coefficient of $x^u z_1^{n_1} z_2^{n_2}$ in this function then dividing by the total number $\rho_{n_1,n_2}$ of instances with given group sizes $n_1$ and~$n_2$ gives the conditional probability that the intersection of the projections has size~$u$, knowing the group sizes:
\begin{equation}
Pr ( |\pi_\Y (G_1) \cap \pi_\Y (G_2)| = u / n_1, n_2) =
\frac{[x^u z_1^{n_1} z_2^{n_2}] F(x; 1,1;0;z_1,z_2)  }{\rho_{n_1,n_2}}.
\label{eq:proba-inter-non-vide}
\end{equation}
The numerator of that expression is
\begin{eqnarray*}
&& [x^u z_1^{n_1} z_2^{n_2}] \left( 1 + \phi (z_1) + \phi(z_2) + x [ \phi(z_1+z_2) - \phi(z_1) - \phi(z_2)  ]  \right)^{d_\Y}
\\ &=&
\binom{d_\Y}{u} [ z_1^{n_1} z_2^{n_2}] \, \left( 1 + \phi (z_1) + \phi(z_2))^{d_\Y -u} \; (\phi(z_1+z_2) - \phi(z_1) - \phi(z_2) \right)^u .
\end{eqnarray*}
In order to simplify it, we first rewrite the polynomial term $\phi(z_1+z_2) - \phi(z_1) - \phi(z_2)$ as  
$(1 + \phi(z_1+z_2)) - (1 + \phi(z_1) + \phi(z_2) )$.
Then we can write its $u$-th power as
\[
 \sum_{0 \leq v \leq u} \binom{u}{v} (-1)^{u-v} ( 1 + \phi (z_1) + \phi(z_2))^{u-v} \; (1+\phi(z_1+z_2))^v.
\]
This gives
\begin{eqnarray*}
&&( 1 + \phi (z_1) + \phi(z_2))^{d_\Y -u} \; (\phi(z_1+z_2) - \phi(z_1) - \phi(z_2) )^u
\\ &&=
 \sum_v \binom{u}{v} (-1)^{u-v} ( 1 + \phi (z_1) + \phi(z_2))^{d_\Y-v} \; (1+\phi(z_1+z_2))^v.
\end{eqnarray*}
Define now
\[
A_{n_1,n_2;v}: = [ z_1^{n_1} z_2^{n_2} ] ( 1 + \phi (z_1) + \phi(z_2))^{d_\Y-v} \; (1+\phi(z_1+z_2))^v .
\]
and write the conditional probability that the intersection of the projections has size~$u$ as
\[
 Pr (| \pi_\Y (G_1) \cap \pi_\Y (G_2)| = u / n_1, n_2)
=
\frac{\binom{d_\Y}{u} \; \sum_{v=0}^u (-1)^{u-v} \, \binom{u}{v} \, A_{n_1,n_2;v}  }{\rho_{n_1,n_2}} .
\]
Our next step is to obtain a manageable expression for the $A_{n_1,n_2;v}$.
Now each product $( 1 + \phi (z_1) + \phi(z_2))^{d_\Y-v} \; (1+\phi(z_1+z_2))^v$ can itself be written as
\begin{eqnarray*}
&& ( \phi (z_1) + (1+z_2)^{d_\Z})^{d_\Y-v} \; (1+z_1+z_2)^{v \, d_\Z}
\\ &&=
\sum_{k \geq 0} \binom{d_\Y-v}{k} \, \phi(z_1)^k  \, (1+z_2)^{d_\Z (d_\Y-v-k)} \;
\sum_{\ell \geq 0} \binom{v \, d_\Z}{\ell} \, z_1^\ell \, (1+z_2)^{v \, d_\Z - \ell}
\\ &&=
\sum_{k, \ell \geq 0} \binom{d_\Y-v}{k}  \binom{v \, d_\Z}{\ell} \; z_1^\ell \, \phi(z_1)^k \; (1+z_2)^{d_\Z (d_\Y-k)-\ell}.
\end{eqnarray*}
The coefficient of $z_1^{n_1} z_2^{n_2}$ in it is
\begin{equation}
    \label{eq:Av}
A_{n_1,n_2;v} =
\sum_{k, \ell \geq 0} \binom{d_\Y-v}{k}  \binom{v \, d_\Z}{\ell}  \; \binom{d_\Z (d_\Y-k)-\ell}{n_2} \; \alpha_{k;n_1-\ell}.
\end{equation}
Summing up, we obtain the probability that the intersection of the projections has size~$u$ as
\begin{equation}
    \label{eq:proba-size-inter}
 Pr (| \pi_\Y (G_1) \cap \pi_\Y (G_2)| = u / n_1, n_2)
=
\frac{\binom{d_\Y}{u} \; \sum_{v=0}^u (-1)^{u-v} \, \binom{u}{v} \, A_{n_1,n_2;v}  }{\rho_{n_1,n_2}} 
\end{equation}
where the $A_{n_1,n_2;v}$ are given by Equation~\eqref{eq:Av}.

\paragraph{Upper bound on the intersection size}
Should we wish to compute the probability that the intersection of the projections has size at most~$t$, we can assume that $t < d_Y$: this probability is obviously equal to 1 for $t = d_\Y$.
 Equation~\eqref{eq:proba-size-inter} readily provides an answer:
\begin{eqnarray*}
 Pr (| \pi_\Y (G_1) \cap \pi_\Y (G_2)|\leq t / n_1, n_2) &=&
 \sum_{u=0}^t  Pr (| \pi_\Y (G_1) \cap \pi_\Y (G_2)| = u)
 \\ &=&
 \frac{\sum_{0 \leq v \leq u \leq t} (-1)^{u-v} \,  \binom{d_\Y}{u} \,\binom{u}{v} \, A_{n_1,n_2;v})  }{\rho_{n_1,n_2}}
 \\ &=&
 \frac{\sum_{v=0}^t \left[ \sum_{u=v}^t (-1)^{u-v} \,  \binom{d_\Y}{u} \,\binom{u}{v} \right] \; \, A_{n_1,n_2;v} . }{\rho_{n_1,n_2}}
\end{eqnarray*}
We can simplify the bracketed sum and write it as $(-1)^{t-v} \binom{d_\Y}{v} \binom{d_\Y-v-1}{t-v}$, hence the final expression for the probability:
\begin{equation}
    \label{eq:proba-size-max-inter}
Pr (| \pi_\Y (G_1) \cap \pi_\Y (G_2)|\leq t / n_1, n_2) =
\frac{\sum_{v=0}^t  (-1)^{t-v} \binom{d_\Y}{v} \binom{d_\Y-v-1}{t-v}  \, A_{n_1,n_2;v} }{\rho_{n_1,n_2}}.
\end{equation}

\paragraph{Numerical computations}
We now have an expression for the general probability that the size of the intersection of the projections has size (at most)~$u$, and efficient computation of the terms $A(v)$ through Equation~\eqref{eq:Av} is clearly the crucial point.

From the definitions of sizes, we know that $n_1, n_2 \geq 1$ and $0 \leq u \leq inf(n_1,n_2,d_\Y)$; furthermore $0 \leq v \leq u$.

The indices in the double sum range a priori from~0 to~$d_\Y-v$ for~$k$ and from~0 to~$v \, d_\Z$ for $\ell$, but can be further reduced as follows.
First, $\alpha_{k;n_1-\ell} \neq 0$ implies that $k + \ell \leq n_1 $, hence $k \leq n_1$, and similarly $\ell \leq n_1-k \leq n_1$.
Then $\binom{d_\Z (d_\Y-k)-\ell}{n_2} \neq 0$ implies that $n_2 \leq d_\Z (d_\Y -k) - \ell$, which we rewrite as $\ell \leq d_\Z (d_\Y -k) - n_2$.
Putting all this together leads to
\begin{eqnarray*}
\begin{cases}
k \leq\min  (d_\Y - v , n_1) ;
\\
\ell  \leq  \min ( v d_\Z, n_1 -k , d_z (d_\Y-k) - n_2   ).
\end{cases}
\end{eqnarray*}
The double sum can thus be reduced to (at worst) $n_1^2$ iterations, i.e. it is quadratic in~$n_1$ in the worst case and quite efficient for the numerical values we consider.

\subsection{Some asymptotic Computations}
\label{sec:calculs-asympt}

Until now, we have been working with the so-called ``ordinary'' generating function $\phi = (1+z)^{d_\Z} - 1$ counting the number of instances of given size, but we might as well use the probability generating function $\phi = (1+z/d_\Z)^{d_\Z} - 1$ (assuming all $\Z$-values to be equally likely): this amounts to introducing a coefficient $d_\Z^n$ in $[ z^n ] \phi(z)$, which disappears when computing conditional probabilities.
We are in the case where $d_\Z$ grows large, in this situation the function $(1+z/d_\Z)^{d_\Z}$ converges towards $e^z$, and we might consider substituting $e^z -1$ to the function $\phi(z)$.
However, such a change corresponds to a change in the kind of generating function we use: we now consider the \emph{exponential} generating function for the number $\phi_n$ of instances of size~$n$ we want to count (see~\cite[Part~A]{livre-flajolet} for a presentation of the types of generating functions and when to use them).
In the rest of this section we set $ $; however this is now 
\[
\phi(z) = \sum_n \phi_n \frac{z^n}{n!}  = e^z -1 ;
\]
accordingly we switch to the notation
\[
\phi_n = \left[  \frac{z^n}{n!}   \right] \phi(z) .
\]

\medskip
To illustrate how the asymptotic approximation of coefficients might work, we consider Case {\bf (A)} of Section~\ref{sec:appendice_differentscas_M1}: the generating function for the number of instances is now
\[
F(1,1;z,z) = \left( 2  e^z -1    \right)^{d_\Y} .
\]
Recall that each value in $D_\Y$ may either not appear in the instance, or appear in exactly one of the two groups of positive and negative observation, each of these cases being associated with the function $e^z -1$.

We assume that {\em $d_\Y$ is fixed} (hopefully a small constant). 
To simplify the notations, we set $d := d_\Y$ and $G(z) := F(1,1;z,z) = \left( 2 \, e^z -1 \right)^{d}$.

Now consider the coefficient $\rho_n = [ z^n / n! ] G(z)$.
We are in the classical framework of computing an asymptotic approximation for the coefficients of an entire function, and it is not difficult to work out a saddle-point approximation for $\rho_n$ (see~\cite[Ch.~VIII]{livre-flajolet}).
However a simpler approach also answers our needs: we expand $\phi(z)$ into the sum of a \emph{fixed} number of terms (recall that $d$ is fixed) to obtain
\begin{eqnarray*}
G(z) = \sum_{i=0}^d \, (-1)^{i}  \, {d \choose i} \, 2^{d-i} \, e^{(d-i) z}.
\end{eqnarray*}
Then
\[
\rho_n = \sum_{i=0}^d \, (-1)^{i} \, 2^{d-i} \, {d \choose i} \, (d-i)^n
.
\]
Asymptotically, the first term is the dominant one and we readily obtain
\[
\rho_n = 2^d \, d^n \, \left( 1 + O \left( ( 1-1/d )^n \right) \right).
\]

\medskip
Consider now the computation of $\rho_{n_1,n_2}$, a simple situation that involves two parameters.
Again we have that
\[
\rho_{n_1,n_2} = \left[ \frac{z^{n_1}}{n_1 !} \frac{z^{n_2}}{n_2 !} \right] G(z_1,z_2)
\qquad {\rm with} \qquad
G(z_1, z_2) = \left( e^{z_1} + e^{z_2} -1 \right)^d.
\]
Taking $d=2$ here corresponds to a set $\Y$ with a single value; then
\begin{eqnarray*}
G(z_1, z_2) &=& \left( e^{z_1} + e^{z_2} -1 \right)^2
\\ &=& 
 e^{2 z_1} + 2 e^{z_1} e^{z_2} + e^{2 z_2} - 2 e^{z_1} - 2 e^{z_2} + 1.
\end{eqnarray*}
The only term that contributes to $\rho_{n_1,n_2}$ is the one in which both $e^{z_1}$ and $ e^{z_2}$ appear, i.e.,  for which the sets of positive and negative observations are both non-empty.
This is the term $2 e^{z_1} e^{z_2}$, which gives $\rho_{n_1,n_2} = 2$ (we just have to decide which of the two possible $Y$ values goes into the set of positive observations).

The next possible value is $d=3$, which we consider even though it is not a power of~2:
\begin{eqnarray*}
G(z_1,z_2) &=&
e^{3 z_1}  \,+\, 3 e^{2 z_1} e^{z_2} \,+\, 3e^{ z_1} e^{2 z_2} \,+\, e^{3 z_2}
\,-\,  3 e^{2 z_1} \,-\, 6 e^{z_1} e^{z_2} 
\\&&
\,-\, 3\, e^{2 z_2}
\,+\, 3 e^{z_1} 
\,+\, 3 e^{z_2} \, - \, 1
\end{eqnarray*}
The terms contributing to $\rho_{n_1,n_2} $ are
$ 3 e^{2 z_1} e^{z_2} \,+\, 3e^{ z_1} e^{2 z_2}  \,-\, 6 e^{z_1} e^{z_2}$
and we obtain
\[
\rho_{n_1,n_2} = 3 \left( 2^{n_1} + 2^{n_2} \right) - 6.
\]
Should we wish to compute, e.g., the ratio $\rho_{n_1,n_2} / \rho_n$, we have to make an assumption on the relative orders of growth of $n_1$ and $n_2$.
Let us assume that they are proportional, i.e., that $n_1 = \alpha \, n$ and $n_2 = (1-\alpha)\, n$ with $0 < \alpha < 1$ a constant.
Here $\rho_n \sim 8 \, 3^n$.
\begin{itemize}
\item[--] 
	The case $\alpha = 1/2$, i.e., $n_1 = n_2$, is specific and we consider it first. 
	We have that $\rho_{n_1,n_2} = 6 (2^{n/2} -1)$, and
	$\rho_{n_1,n_2} / \rho_n \sim \frac{3}{4} \, 2^{n/2} \, 3^{-n}  $.
	
\item[--]
	Let us now assume that $\alpha > 1/2$ (the case $\alpha < 1/2$ is symmetrical).
	Then the dominant term in $\rho_{n_1,_2}$ is $2^{n_1} = 2^{\alpha n}$ and the ratio $\rho_{n_1,n_2} / \rho_n$ becomes
	$ \frac{3}{8} \, (2^\alpha / 3)^n = \frac{3}{8} \, e^{-\beta n} $ with $\beta = \log 3 - \alpha \log 2 > 0$.
	
\end{itemize}

\medskip
A similar computation may be done for any value of~$d$; we give below the result for $d=4$ corresponding to the situation where the set $\Y$ has two elements:
\[
\rho_{n_1,n_2} = 4 \left( 3^{n_1} + 3^{n_2} \right) + 6 \, 2^{n_1 + n_2} - 12 \left( 2^{n_1} + 2^{n_2} -1 \right).
\]
Again, to proceed further we need to know the respective orders of growth of $n_1$ and $n_2$.

Two things should now be clear:
\begin{itemize}

\item[--] the information we need in this multivariate problem, namely the respective orders of growths of the various sizes we consider, can only come from knowledge of the exact problem we are modelizing; 

\item[--] once we have this information, we are in a position to compute asymptotic expressions for $\rho_{n_1,n_2}$, or indeed any coefficient we might be interested in.

\end{itemize}

\bibliography{biblio}

\end{document}

%% file: commandes.tex
\newcommand{\C}{{\cal C}}

\newcommand{\X}{{\cal X}}

\newcommand{\Y}{{\cal Y}}
\newcommand{\Z}{{\cal Z}}

\newcommand{\True}{{\sf True}}
\newcommand{\False}{{\sf False}}
\newcommand{\MCP}{{\sf MCP}}
\newcommand{\Patterns}{{\sf Patterns}}

\DeclareRobustCommand{\stirling}{\genfrac\{\}{0pt}{}}

\newtheorem{definition}{\bf Definition}
\newtheorem{property}{\bf Property}
\newtheorem{example}{\bf Example}